\documentclass[10pt,twocolumn,letterpaper]{article}

\usepackage{cvpr}
\usepackage{times}
\usepackage{epsfig}
\usepackage{graphicx}
\usepackage{amsmath}
\usepackage{amssymb}

\usepackage{algorithm}
\usepackage{algorithmic}
\usepackage{amsfonts}
\usepackage{amsthm}  
\usepackage{wasysym}
\usepackage[ruled,vlined,algo2e]{algorithm2e}
\providecommand{\SetAlgoLined}{\SetLine}
\usepackage{color}
\usepackage{dsfont}	
\usepackage{wrapfig}
\usepackage{footnote}
\usepackage{epstopdf}

\theoremstyle{definition}

\DeclareMathOperator*{\argmin}{arg\,min}

\newcommand{\resp}{{\it resp. }}

\newcommand{\rom}[1]{\lowercase\expandafter{\romannumeral #1\relax}}

\usepackage[pagebackref=true,breaklinks=true,letterpaper=true,colorlinks,bookmarks=false]{hyperref}

\cvprfinalcopy 

\def\cvprPaperID{1725} 

\ifcvprfinal\fi
\begin{document}

\title{Efficient Training of Very Deep Neural Networks for Supervised Hashing}

\author{Ziming Zhang, Yuting Chen and Venkatesh Saligrama\\
Department of ECE and Division of Systems Engineering, Boston University\\
{\tt\small \{zzhang14, yutingch, srv\}@bu.edu}
}

\maketitle

\begin{abstract}
In this paper, we propose training very deep neural networks (DNNs) for supervised learning of hash codes. Existing methods in this context train relatively ``shallow'' networks limited by the issues arising in back propagation (\eg vanishing gradients) as well as computational efficiency. We propose a novel and efficient training algorithm inspired by alternating direction method of multipliers (ADMM) that overcomes some of these limitations. Our method decomposes the training process into independent layer-wise local updates through auxiliary variables. 
%
Empirically we observe that our training algorithm always converges and its computational complexity is linearly proportional to the number of edges in the networks. 
Empirically we manage to train DNNs with 64 hidden layers and 1024 nodes per layer for supervised hashing in about 3 hours using a single GPU. Our proposed very deep supervised hashing (VDSH) method significantly outperforms the state-of-the-art on several benchmark datasets.
\end{abstract}


\section{Introduction}
Supervised hashing techniques aim to learn compact and similarity-preserving binary representations from labeled data, such that similar inputs are mapped to nearby binary hash codes in the Hamming space, and information
retrieval can be efficiently and effectively done in large-scale databases. A large category of these methods seek to learn a set of hyperplanes as linear hash functions, such as Iterative Quantization (ITQ) \cite{gong2013iterative}, supervised Minimal Loss Hashing (MLH) \cite{norouzi2011minimal},  Semi-Supervised Hashing (SSH) \cite{wang2012semi}, and FastHash \cite{lin2014fast}. Several kernel-based hashing methods like Binary Reconstructive Embedding (BRE) \cite{kulis2009learning} and Kernel-Based Supervised Hashing (KSH) \cite{liu2012supervised} have also been proposed.

It is well recognized that deep models are able to learn powerful image representations in a latent space where samples with different properties can be well separated.
In this context convolutional Neural Networks (CNN) based hashing schemes have been developed \cite{Erin_2015_CVPR, conf/icdm/KangKC12, conf/esann/KrizhevskyH11, Wang:2015:DMH:2832415.2832567, xia2014supervised, DBLP:journals/corr/ZhangLZZZ15, zhao2015deep}. Hash codes learned from these latent spaces have been shown to significantly improve the retrieval performance on many benchmark datasets. 



Nevertheless, the efficacy of deep learning in applications such as hashing hinges on the ability to efficiently train deep models \cite{glorot2010understanding}. 
Back propagation (or ``backprop'') \cite{Russell:2003:AIM:773294} is currently the most widely-used training method in deep learning due to its simplicity. Backprop is known to suffer from the so called vanishing gradient issue \cite{hochreiter2001gradient}, where gradients in the front layers of an $n$-layer network decrease exponentially with $n$. This directly impacts computational efficiency, which in turn limits the size of the networks that can be trained.
%
For instance, the training of VGG's very deep features \cite{simonyan2014very} for ILSVRC2014 with 16 convolutional layers takes approximately one month using 4 GPUs.


\noindent
{\bf Contributions:} We propose a {\em very deep supervised hashing (VDSH)} algorithm by training very deep neural networks for hashing. 
Our method can take in any form of vector input, such as raw image intensities, traditional features like GIST \cite{oliva2001modeling}, or even CNN features \cite{lecun2013lenet}. Given training data with class labels, our network learns a data representation tailored for hashing, and outputs binary hash codes with varying lengths. VDSH can easily train large very deep networks within hours on a single GPU.

Our learning objective is to generate optimal hash codes for linear classification. To this end we minimize the least square between the weighted encoding features (\ie the output of our last hidden layer) and their label vectors with regularization on model parameters to prevent overfitting.


Rather than using backprop, we propose a novel computationally efficient training algorithm for VDSH inspired by alternating direction method of multipliers (ADMM) \cite{boyd2011distributed}. We represent DNN features in a recursive way by introducing an auxiliary variable to model the output of each hidden layer for each data sample. 
Then we apply the augmented Lagrangian to incorporate our learning objective with equality constraints, where another set of auxiliary variables are introduced to store the network weights between every pair of adjacent layers locally for efficient update. 


Empirically we demonstrate smooth convergence and computational efficiency for VDSH. Our training complexity is linearly proportional to the number of connections between nodes in the network. 
We train DNNs with up to 64 hidden layers and 1024 nodes per layer for supervised learning of hash codes within about 3 hours on a single GTX TITAN GPU, while achieving state-of-the-art results on several benchmark datasets.

\subsection{Related Work}
\noindent
{\bf (\rom{1}) Supervised hashing with deep models:} 
Learning high-level feature representations by building deep hierarchical models have shown great potential in various applications. Researchers have been adopting deep models to jointly learn image representations and hash codes from data. Kang \etal \cite{conf/icdm/KangKC12} proposed a deep multi-view hashing (DMVH) algorithm to learn hash codes with multiple data representations. Xia \etal \cite{xia2014supervised} proposed learning image representations for supervised hashing by approximating the data affinity matrix with CNN features. Zhao \etal \cite{zhao2015deep} proposed a Deep Semantic Ranking Hashing (DSRH) method to preserve multilevel semantic similarity between multi-label images. Erin Liong \etal \cite{Erin_2015_CVPR} proposed a deep hashing method to explore the nonlinear relationships among data. Zhang \etal \cite{DBLP:journals/corr/ZhangLZZZ15} proposed a Deep Regularized Similarity Comparison Hashing (DRSCH) method to allow the length of output bits to be scalable. Most of the works learn hash functions on top of a deep CNN architecture. In contrast, VDSH can be built from arbitrary vector representations. When CNN features are used, our method can be viewed as fine-tuning these networks for supervised hashing. Besides, the scale and depth of our DNNs are much larger than previous methods, which pose harder challenges for training.


\noindent
{\bf (\rom{2}) DNNs:} In the literature, many different DNN architectures (\eg LeNet \cite{lecun2013lenet}, AlexNet \cite{krizhevsky2012imagenet}, GoogLeNet \cite{szegedy2014going} and VGG-VD \cite{simonyan2014very}) and weighting structures (\eg sparse network \cite{ChengICCV2015}, circulant structure \cite{liu2015sparse}, low-rank approximation \cite{sainath2013low}) have been proposed. Several techniques have been proposed to improve the generalization of networks such as dropout \cite{srivastava2014dropout} and dropconnet \cite{wan2013regularization}, which can be viewed as better regularization. Some techniques for speeding-up the training have been proposed as well such as distributed training \cite{dean2012large} and batch normalization \cite{ioffe2015batch}. These architectures and methods, however, are trained using backprop, suffering from the same issues such as vanishing gradients.

Ongoing efforts to overcome issues in backprop include variational Bayesian autoencoder \cite{kingma2013auto}, auto-encoding target propagation \cite{bengio2015towards}, and difference target propagation \cite{lee2015difference}. 

Carreira-Perpin{\'a}n and Wang \cite{carreira2014distributed} recently proposed a method for training deeply nested systems. Their method of auxiliary coordinates (MAC) breaks down the dependency in nested systems into equality constraints, so that the quadratic penalty method can be utilized as an efficient solver. Shen \etal \cite{shen2015supervised} proposed a Supervised Discrete Hashing (SDH) method based on MAC which achieved the state-of-the-art on supervised hashing. Carreira-Perpin{\'a}n and Raziperchikolaei \cite{carreira2015hashing} proposed learning binary autoencoders for hashing as well using MAC. 

In contrast our ADMM-based method is more suitable and efficient for solving regularized loss minimization as has been shown in the Block-Splitting algorithm \cite{parikh2014block}. ADMM solves optimization (possibly nonconvex) problems with equality constraints by decomposing an objective into several disjoint sub-objectives using new auxiliary variables so that the original objective can be optimized iteratively using coordinate descent. With small additional computational cost we circumvent the need for relaxation of penalty related parameters as required in this context \cite{GVK502988711}.

\section{Very Deep Supervised Hashing}

Our problem setup closely mirrors \cite{shen2015supervised}. We are given a collection of $N$ samples $\mathcal{X} =  \{ \mathbf{x}_i  \}_{i=1}^N \in \mathbb{R}^{d\times N}$. Our goal is to learn a collection of K-bit binary codes $\mathbf{B} \in \{-1,1\}^{K \times N}$ where the $i$-th column $b_i \in \{-1,1\}^K$ denotes the binary code for the $i$-th sample $\mathbf{x}_i$. 

To learn these codes we consider a parameterized family of models, $F(\mathbf{x},\mathbf{\Theta})$, parameterized by $\mathbf{\Theta}$, that map an arbitrary element $\mathbf{x} \in \mathcal{X}$ to $\mathbb{R}^K$. The hash code for a particular model described by $\mathbf{\Theta}$ is then obtained by taking the sign of $F$, namely,
\begin{align}\label{eqn:hashing}
\mathbf{b}_i = \mbox{sgn}(F(\mathbf{x}_i,\mathbf{\Theta})),
\end{align}
where sgn denotes the entry-wise sign function,  \ie $\mbox{sgn}(x)=1$ if $x>0$, otherwise $\mbox{sgn}(x)=-1$.

In supervised hashing we are also provided with class labels for the $N$ samples and the goal in this context is to ensure that the binary codes for the samples corresponding to each class are similar. We adopt the perspective of \cite{shen2015supervised} in that 
binary codes that are learned in the context of linear classification are good hashing codes, namely, they preserve semantic similarity of the data samples. To this end, we encode the ground truth for each of the $C$ classes into $C$-dim binary vectors, $\mathbf{y}_i,\,i=1,\ldots,N$ where the $j$-th entry $y_{ji}=1$ if $\mathbf{x}_i$ belongs to class $j$. Our hypothesis suggests that there is a collection of $C$ linear classifier functions, $\mathbf{w}_1,\mathbf{w}_2,\ldots,\mathbf{w}_C$ such that the predicted output
$
\hat{\bf y}_i = [\mathbf{w}_1^T\mathbf{b}_i,\, \mathbf{w}_2^T\mathbf{b}_i,\, \ldots,\, \mathbf{w}^T_C\mathbf{b}_i]^T=\mathbf{W}^T\mathbf{b}_i
$
closely matches the ground-truth label vector $\mathbf{y}_i$ for data $\mathbf{x}_i$, where $(\cdot)^T$ denotes the matrix transpose operator. In other words, we seek hash codes and linear classifiers $\mathbf{W}$ such that $\hat {\bf y}_i \approx \mathbf{y}_i$, where the approximation error is measured in terms of some loss function $L$. This leads to the following optimization problem as in \cite{shen2015supervised}:
\begin{align}
\min_{\boldsymbol{\Theta}, \mathbf{W},\mathbf{B}} & \sum_i 
L(  \mathbf{W}^T \mathbf{b}_i , \mathbf{y}_i )+ \Omega(\boldsymbol{\Theta}, \mathbf{W}), \\ \nonumber
& \mbox{s.t.} \,\,\,\, \mathbf{b}_i = \mbox{sgn}(F(\mathbf{x}_i, \boldsymbol{\Theta})), \, \forall i.
\end{align}
Note that this formulation is identical to an unconstrained objective function, namely,
\begin{align}
\min_{\boldsymbol{\Theta}, \mathbf{W}} \sum_i 
L(  \mathbf{W}^T \mbox{sgn}(F(\mathbf{x}_i, \boldsymbol{\Theta})) , \mathbf{y}_i )+ \Omega(\boldsymbol{\Theta}, \mathbf{W}).
\end{align}
Much of the difficulty arises from the need to deal with the sign function. A number of researchers (see \cite{shen2015supervised}) have proposed various techniques to deal with this problem. These include (a) approximating the sign function using sigmoids (\eg \cite{liu2012supervised}); (b) penalizing deviations between $F(\cdot,\mathbf{\Theta})$ and $\mathbf{B}$ (\eg \cite{shen2015supervised}); (c) relaxing the binary constraint to be continuous (\eg \cite{wang2012semi}), \ie $\mathbf{b}_i = F(\mathbf{x}_i, \mathbf{\Theta})$. We adopt approach (c), where we first learn the continuous embeddings
$\mathbf{b}_i$ and then threshold them later to be binary codes. This leads to our objective in training VDSH as follows: 
\begin{align}\label{eqn:obj}
\min_{\boldsymbol{\Theta}, \mathbf{W}} \sum_i 
L(  \mathbf{W}^T F(\mathbf{x}_i, \boldsymbol{\Theta}) , \mathbf{y}_i )+ \Omega(\boldsymbol{\Theta}, \mathbf{W}).
\end{align}
While \cite{shen2015supervised} suggests that this method can be fast, it may lead to sub-optimal performance. As we will see in our experiments this potential suboptimality is offset by training very deep models resulting in significantly better performance relative to \cite{shen2015supervised}. 
For simplicity, we choose squared loss functions and penalties (although many other choices such as hinge loss, $\ell_1$ norm penalty \etc are all possible). Specifically, we let $L(  \mathbf{W}^T F(\mathbf{x}_i, \boldsymbol{\Theta}) , \mathbf{y}_i ) = \frac{1}{2}\left\|\mathbf{W}^T F (\mathbf{x}_i, \boldsymbol{\Theta}) - \mathbf{y}_i\right\|_2^2$ be a square loss function. $\Omega(\boldsymbol{\Theta}, \mathbf{W})=\frac{\alpha_{\theta}}{2}\sum_m\|\boldsymbol{\theta}^{(m)}\|_2^2 + \frac{\alpha_W}{2}\|\mathbf{W}\|_F^2$ denotes a joint regularizer over $\boldsymbol{\Theta}$ and $\mathbf{W}$, where $\|\cdot\|_2$ and $\|\cdot\|_F$ denote $\ell_2$ norm and Frobenius norm, respectively, and $\alpha_{\theta}\geq0$ and $\alpha_{W}\geq0$ are regularization parameters.

\subsection{Very Deep Hashing Model}\label{ssec:hash}
We formally describe our parameterized model for $F(\mathbf{x},\mathbf{\Theta})$ in this section. Our very deep hashing model (VDSH) is a network with $M$ hidden layers given by:
\begin{align}\label{eqn:F}
\begin{split}
& F_0(\mathbf{x}_i)  = \mathbf{x}_i, \\
& F_{m}(\mathbf{x}_i)  = f_{m}(F_{m-1}(\mathbf{x}_i); \boldsymbol \theta^{(m)}), 1 \leq m \leq M 
\end{split}
\end{align}
where $\boldsymbol{\Theta}=\{ \boldsymbol{\theta}^{(m)} \}_{m=1}^M$ denotes the set of weights for the entire network, each $ \boldsymbol{\theta}^{(m)}\in\mathbb{R}^{D_m\times D_{m-1}} (D_0 = d, D_M = K)$ denotes the weights between the $(m-1)$-th and $m$-th hidden layers, each  $f_m: \mathbb{R}^{D_{m-1}}\mapsto\mathbb{R}^{D_m}$ denotes a nonlinear function which maps the outputs from lower layers $F_{m-1}(\mathbf{x}_i)$ to the outputs of upper layers $F_m(\mathbf{x}_i)$. We let the final layer be    $F(\mathbf{x}_i, \boldsymbol{\Theta}) = F_M(\mathbf{x}_i)$. 
In VDSH we utilize the ReLU \cite{he2015delving} activation function as $f$. In particular, 
\begin{align}\label{eqn:f}
f_m(\mathbf{x}_i; \boldsymbol{\theta}^{(m)}) = \max\left\{\mathbf{0}, \boldsymbol{\theta}^{(m)}\mathbf{x}_i\right\},
\end{align}
where $\max$ is an entry-wise maximum operator. Note that it is possible for our method to incorporate more complex functions to define $f$ so that more complex operations on the hidden nodes can be involved as well, \eg maxout \cite{goodfellow2013maxout}, dropout \cite{srivastava2014dropout}, dropconnet \cite{wan2013regularization}, batch normalization \cite{ioffe2015batch}, and network pruning \cite{han2015learning}. But this discussion is out of the scope of our paper, and we consider it as our future work.

\begin{figure}[t]
\centering
\includegraphics[width = .85\linewidth]{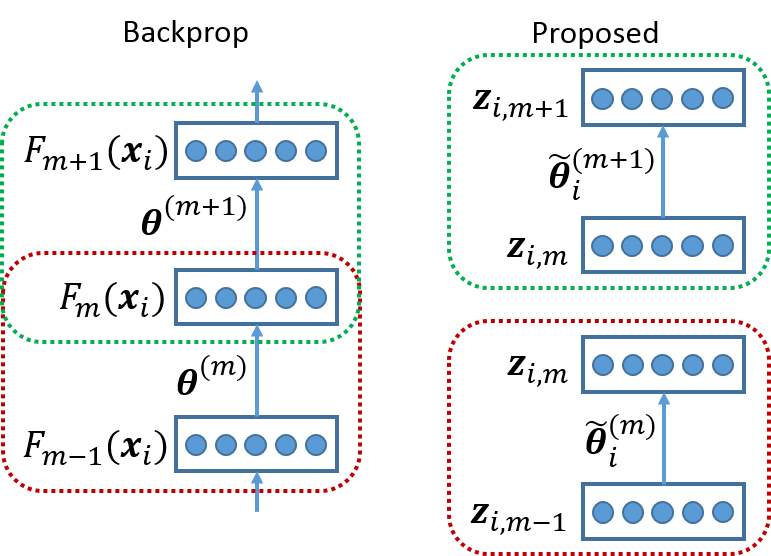}
\vspace{1mm}
\caption{\footnotesize{Schematics of VDSH training algorithm. Blue color represents the network structures, the red and green dashed rectangles represent two two-layer substructures. {\bf (Left)} $F_{m}(\mathbf{x}_i)$ (\resp $F_{m-1}(\mathbf{x}_i)$ and $F_{m+1}(\mathbf{x}_i)$) denotes the output from the $m$-th (\resp $(m-1)$-th and $(m+1)$-th) hidden layers for a data sample $\mathbf{x}_i$. {\bf (Right)} For each data sample we introduce two types of auxiliary variables $\mathbf{z}$ and $\tilde{\boldsymbol{\theta}}$ to represent the outputs of each hidden layer for the data samples and the local copies of network weights for the substructures. Learning the network weights decomposes into independent local learning of weights, leading to efficiency and feasibility of very deep learning}}
\label{fig:targetprop}
\vspace{-5mm}
\end{figure}

\begin{figure*}[t]
\begin{minipage}[b]{0.245\linewidth}
 \begin{center}
 \centerline{\includegraphics[width=\columnwidth]{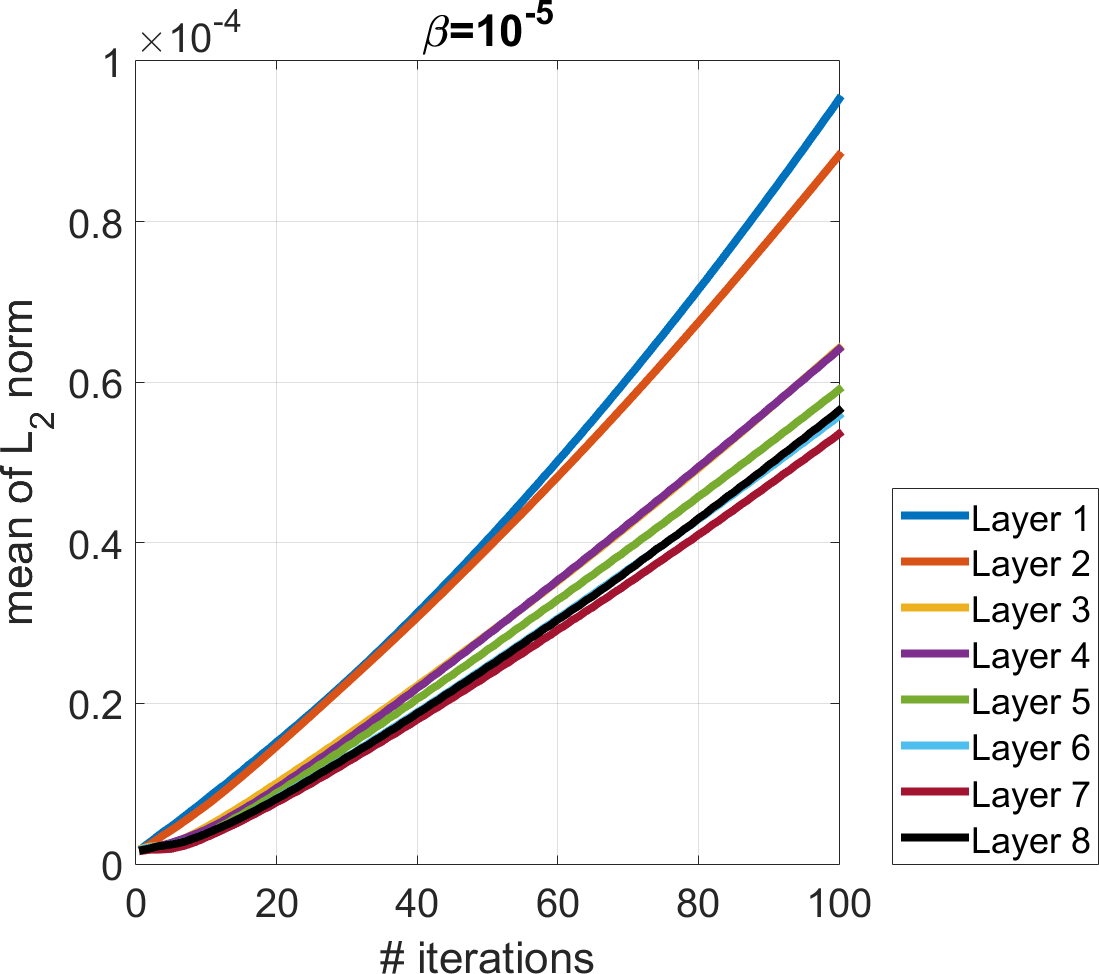}}
 \end{center}
\end{minipage}
\begin{minipage}[b]{0.245\linewidth}
\begin{center}
\centerline{\includegraphics[width=\columnwidth]{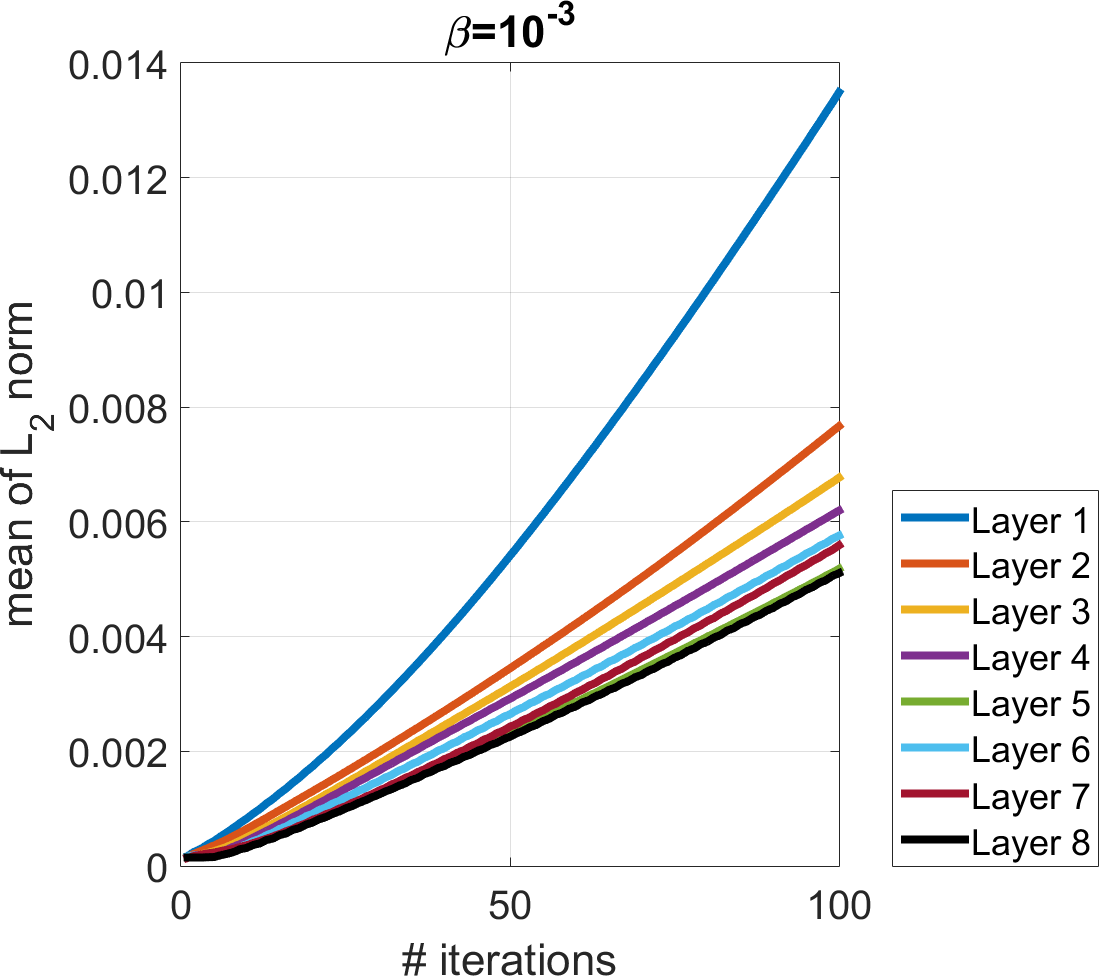}}
\end{center} 
\end{minipage}
\begin{minipage}[b]{0.245\linewidth}
 \begin{center}
 \centerline{\includegraphics[width=\columnwidth]{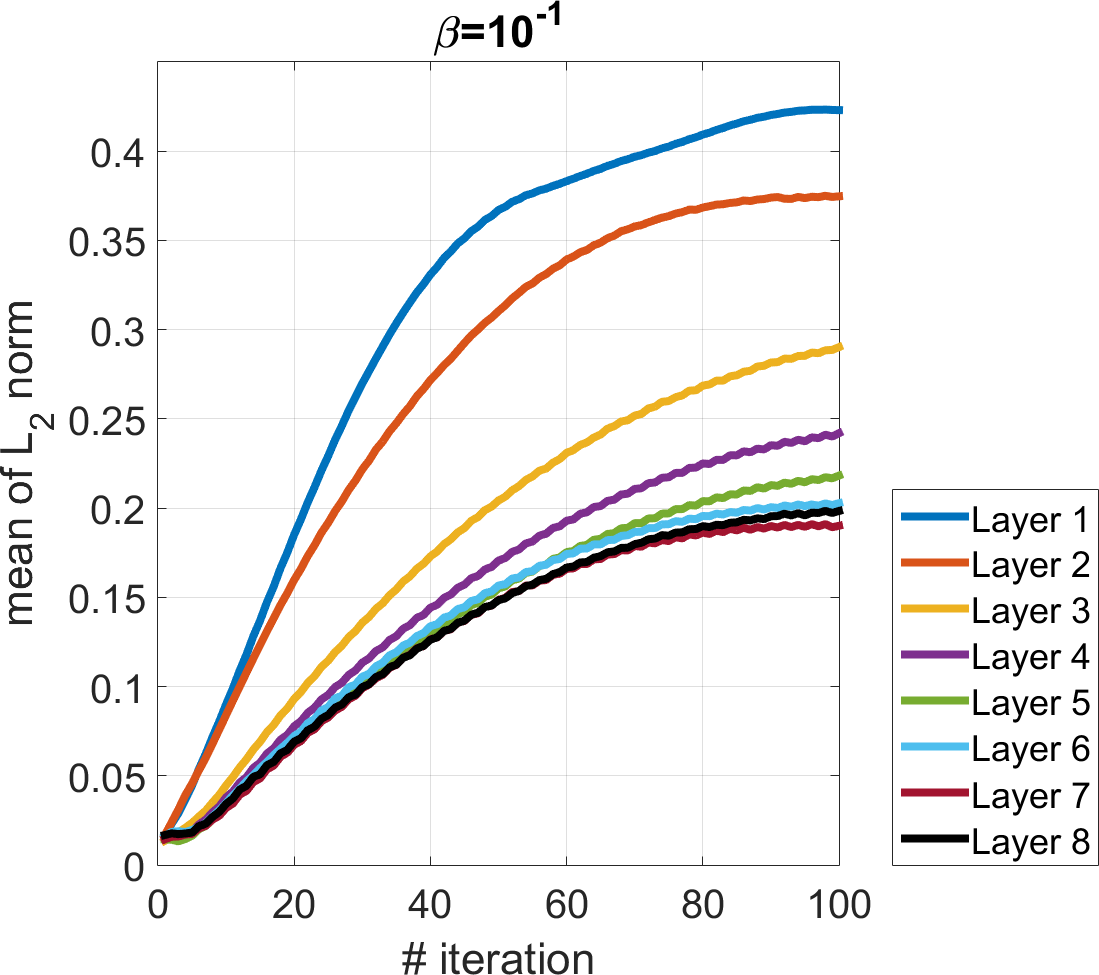}}
 \end{center}
\end{minipage}
\begin{minipage}[b]{0.245\linewidth}
\begin{center}
\centerline{\includegraphics[width=\columnwidth]{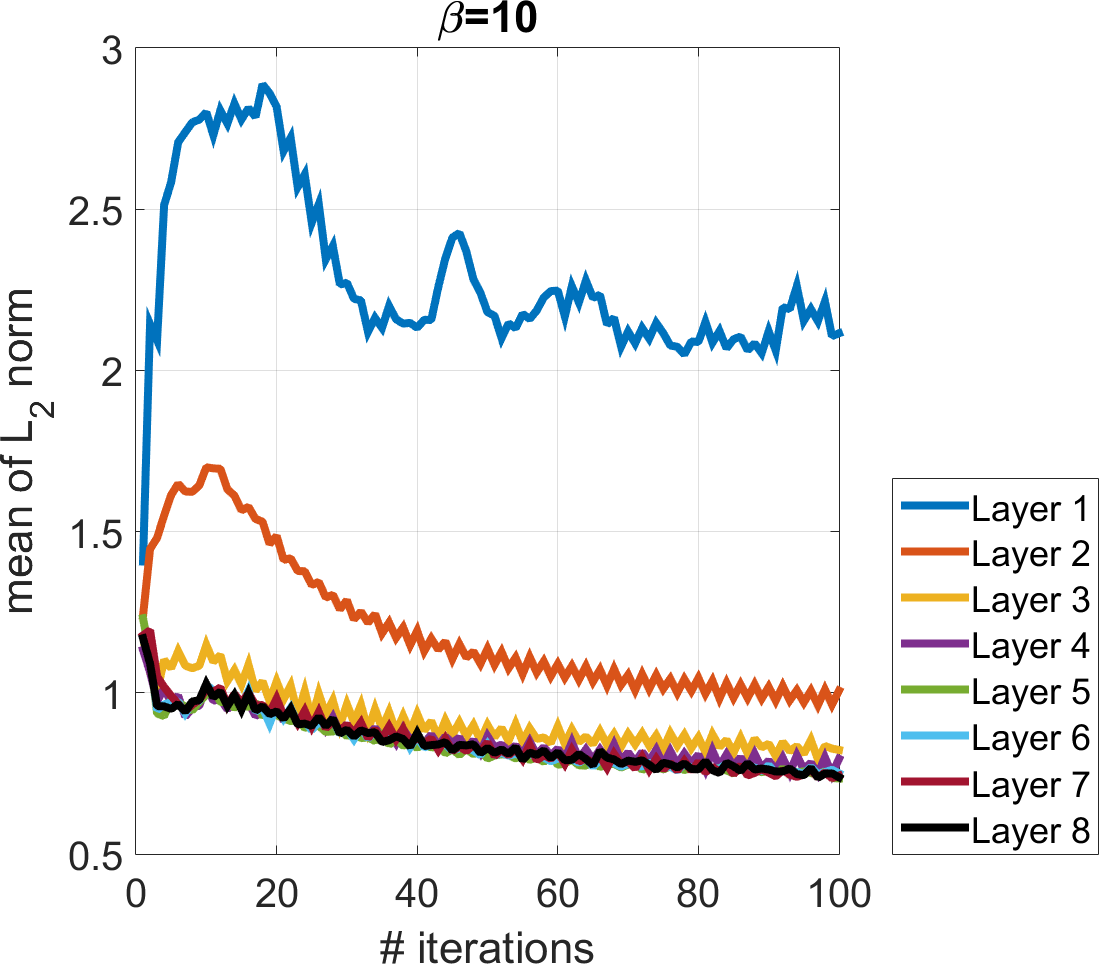}}
\end{center} 
\end{minipage}
\vspace{-3mm}
\caption{\footnotesize{Illustration of empirical convergence of VDSH using the Lagrangian dual variables for auxiliary variables $\mathbf{z}$'s with different dual update steps $\beta$.
}}
\label{fig:convergence}
\vspace{-3mm}
\end{figure*}

\subsection{Optimization}\label{ssec:opt}

While backprop is an option for training VDSH and has been used before for learning hash codes \cite{Erin_2015_CVPR}, it suffers from the well-known ``vanishing gradient problem'' \cite{hochreiter2001gradient} where gradients in the front layers of an $n$-layer network decrease exponentially with $n$. This directly impacts computational efficiency, which in turn limits the size of the networks that can be trained. 
To overcome this problem, we explicitly introduce a set of auxiliary variables $\{\mathbf{z}_{i,m}\}$ for every $\mathbf{x}_i$ at every layer $m$ to represent our network in Eq.~\ref{eqn:F} to circumvent long-term dependencies during training:
%
%
\begin{align}
\mathbf{z}_{i,m} = F_m(\mathbf{x}_i), \; \forall i, \forall 0\leq m\leq M.
\end{align}
In this way, as observed by \cite{carreira2014distributed}, the auxiliary variables break down the network into a collection of two-adjacent-layer substructures (see Fig. \ref{fig:targetprop}).
 
The issue is that we are still left with dependency between the loss function $L$ and the regularizer $\Omega$ (see Fig.~\ref{fig:targetprop}). To circumvent this issue we 
we introduce new auxiliary variables $ \tilde{\boldsymbol{\theta}}_i^{(m)}=\boldsymbol{\theta}^{(m)},\,\,\forall i, \forall m,$ motivated by the block splitting algorithm \cite{parikh2014block}. 
We are now in a position to update network weights $\boldsymbol{\Theta}$ locally and independently across the different layers, which leads to improved computational efficiency. 
We rewrite our objective in terms of these auxiliary variables as follows:
\begin{align}\label{eqn:auxiliary}
& \min_{\boldsymbol{\Theta}, \mathbf{W}, \mathcal{Z}, \tilde{\boldsymbol{\Theta}}} \frac{1}{2}\sum_i\left\|\mathbf{W}^T\mathbf{z}_{i,M} - \mathbf{y}_i\right\|_2^2 + \Omega(\boldsymbol{\Theta}, \mathbf{W}), \\
& \mbox{s.t.} \, \tilde{\boldsymbol{\theta}}_i^{(m)} = \boldsymbol{\theta}^{(m)}, \mathbf{z}_{i,m} = f(\mathbf{z}_{i,m-1}; \tilde{\boldsymbol{\theta}}_i^{(m)}), \forall i, \forall m\in[1, M], \nonumber
\end{align}
where $\mathcal{Z}=\{\mathbf{z}_{i,m}\}$ and $\tilde{\boldsymbol{\Theta}}=\{\tilde{\boldsymbol{\theta}}_i^{(m)}\}$.  Note that unlike conventional ADMM methods the second equality constraint is nonlinear. Our next step is to introduce the augmented Lagrangian as follows:
\begin{align}\label{eqn:admm}
\min_{\boldsymbol{\Theta}, \mathbf{W}, \mathcal{Z}, \tilde{\boldsymbol{\Theta}}, \mathcal{U}, \mathcal{V}} & \frac{1}{2}\sum_i\left\|\mathbf{W}^T\mathbf{z}_{i,M} - \mathbf{y}_i\right\|_2^2 + \Omega(\boldsymbol{\Theta}, \mathbf{W}) \\
& + \frac{\beta}{2}\sum_{i,m}\left\|\mathbf{z}_{i,m} - f(\mathbf{z}_{i,m-1}; \tilde{\boldsymbol{\theta}}_i^{(m)}) + \mathbf{u}_{i,m}\right\|_2^2 \nonumber\\
& + \frac{\gamma}{2}\sum_{i,m}\left\|\boldsymbol{\theta}^{(m)}-\tilde{\boldsymbol{\theta}}_i^{(m)}+\mathbf{v}_{i,m}\right\|_2^2, \nonumber
\end{align}
where $\mathcal{U}=\{\mathbf{u}_{i,m}\}$ and $\mathcal{V}=\{\mathbf{v}_{i,m}\}$ denote the Lagrangian related parameters, $\beta\geq0$ and $\gamma\geq0$ are predefined dual update steps. Note that the Lagrangian dual variables for $\mathbf{z}$'s and $\boldsymbol{\theta}$'s are computed using $ \beta\mathbf{u}_{i,m}$ and $\gamma\mathbf{v}_{i,m},\, \forall i, \forall m,$.

To solve Eq. \ref{eqn:admm}, we propose a novel 
algorithm listed in Alg. \ref{alg:cd}, where $N$ denotes the total number of training samples and $\mathbf{I}$ denotes the identity matrix. For better exposition in Alg. \ref{alg:cd}, we denote $\forall i, \forall m, \mathcal{G}_{i,m}(\cdot)=\left\|\mathbf{z}_{i,m} - f(\mathbf{z}_{i,m-1}; \tilde{\boldsymbol{\theta}}_i^{(m)}) + \mathbf{u}_{i,m}\right\|_2^2$,  $\mathcal{Q}_{i,m}(\cdot)=\left\|\boldsymbol{\theta}^{(m)}-\tilde{\boldsymbol{\theta}}_i^{(m)}+\mathbf{v}_{i,m}\right\|_2^2$. 
We alternatively optimize one variable of $\mathcal{G}$ or $\mathcal{Q}$ at a time.
In each iteration, using the auxiliary variables $\mathbf{z}$'s the classification error is first propagated to the last (or top) hidden layer and then sequentially propagated to the rest of the hidden layers. Next given these updated $\mathbf{z}$'s, the local copies of network weights $\tilde{\boldsymbol{\theta}}_i^{(m)}$ are updated independently. This later leads to updates of the entire network weights $\boldsymbol{\Theta}$. Finally the classifier $\mathbf{W}$ is updated to minimize the total regularized loss while fixing the rest of the parameters. We repeat the updating until the algorithm satisfies convergence condition. Note that since {\it foreach} loop in Alg. \ref{alg:cd} can be updated independently it is amenable to distributed or parallel computation \cite{wang2014parallel}. Nevertheless, we do not pursue it here.

During testing, we utilize the learned network weights $\boldsymbol{\Theta}$ and apply Eq. \ref{eqn:hashing} and \ref{eqn:F} to compute the hash codes.


\begin{algorithm}[t]\footnotesize
\SetAlgoLined
\SetKwInOut{Input}{Input}\SetKwInOut{Output}{Output}
\Input{training data $\{(\mathbf{x}_i, \mathbf{y}_i)\}$ and parameters $\alpha_{\theta}, \alpha_W, \beta, \gamma$}
\Output{network weights $\boldsymbol{\Theta}$}
\BlankLine
Randomly initialize $\boldsymbol{\Theta}, \mathbf{W}$;\\
$\forall i, \forall m\in[1, M], \; \tilde{\boldsymbol{\theta}}_i^{(m)}\leftarrow\boldsymbol{\theta}^{(m)}, \mathbf{v}_{i,m}\leftarrow\mathbf{0}, \mathbf{z}_{i,0}\leftarrow\mathbf{x}_i, \mathbf{z}_{i,m}\leftarrow f(\mathbf{z}_{i,m-1};\tilde{\boldsymbol{\theta}}_i^{(m)}), \mathbf{u}_{i,m}\leftarrow\mathbf{0}$; \\
\Repeat{converge}{
\ForEach{i}{
$\mathbf{z}_{i,M}\leftarrow\argmin_{\mathbf{z}_{i,M}}\hspace{-1mm}\left\{\frac{1}{2}\left\|\mathbf{W}^T\mathbf{z}_{i,M} - \mathbf{y}_i\right\|_2^2 + \frac{\beta}{2}\mathcal{G}_{i,M}(\mathbf{z}_{i,M})\right\}$;\\
$ \mathbf{u}_{i,M}\leftarrow\mathbf{u}_{i,M}+\mathbf{z}_{i,M}-f(\mathbf{z}_{i,M-1}; \tilde{\boldsymbol{\theta}}_i^{(M)})$;
}
\For{$m=M-1:-1:1$}{
$ \forall i, \mathbf{z}_{i,m}\leftarrow\argmin_{\mathbf{z}_{i,m}}\left\{\mathcal{G}_{i,m}(\mathbf{z}_{i,m})+\mathcal{G}_{i,m+1}(\mathbf{z}_{i,m})\right\}$; \\
$ \forall i, \mathbf{u}_{i,m}\leftarrow\mathbf{u}_{i,m}+\mathbf{z}_{i,m}-f(\mathbf{z}_{i,m-1}; \tilde{\boldsymbol{\theta}}_i^{(m)})$;
}
\ForEach{$m$}{
$\forall i, \tilde{\boldsymbol{\theta}}_i^{(m)}\leftarrow\argmin_{\tilde{\boldsymbol{\theta}}_i^{(m)}}\left\{\beta\mathcal{G}_{i,m}(\tilde{\boldsymbol{\theta}}_i^{(m)})+\gamma\mathcal{Q}_{i,m}(\tilde{\boldsymbol{\theta}}_i^{(m)})\right\}$;
$\boldsymbol{\theta}^{(m)}\leftarrow\frac{\gamma}{\gamma N+\alpha_{\theta}}\sum_i\left(\tilde{\boldsymbol{\theta}}_i^{(m)}-\mathbf{v}_{i,m}\right)$;\\
$\forall i, \mathbf{v}_{i,m}\leftarrow\mathbf{v}_{i,m}+\boldsymbol{\theta}^{(m)}-\tilde{\boldsymbol{\theta}}_i^{(m)}$;
}
$\mathbf{W}\leftarrow\argmin_{\mathbf{W}}\hspace{-1mm}\left\{\frac{\alpha_W}{2}\|\mathbf{W}\|_F^2 + \frac{1}{2}\sum_i\left\|\mathbf{W}^T\mathbf{z}_{i,M} - \mathbf{y}_i\right\|_2^2\right\}$;
}
\Return $\boldsymbol{\Theta}$;
\caption{VDSH training algorithm}\label{alg:cd}
\end{algorithm}

\begin{figure*}[t]
\begin{minipage}[b]{0.19\linewidth}
\begin{center}
\centerline{\includegraphics[width=\columnwidth]{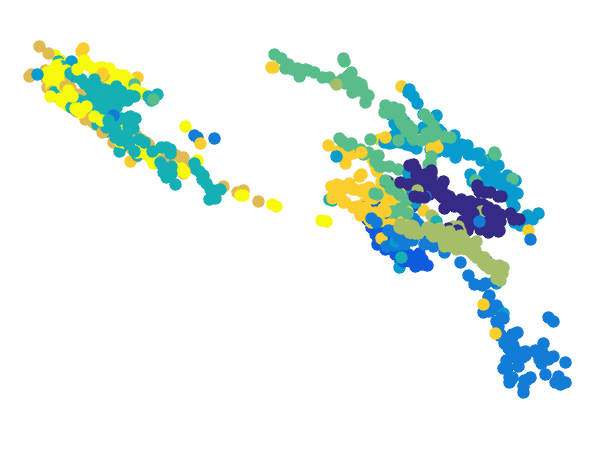}}
\centerline{\footnotesize{(a) Original raw pixel features}}
\end{center}
\end{minipage}
\begin{minipage}[b]{0.19\linewidth}
\begin{center}
\centerline{\includegraphics[width=\columnwidth]{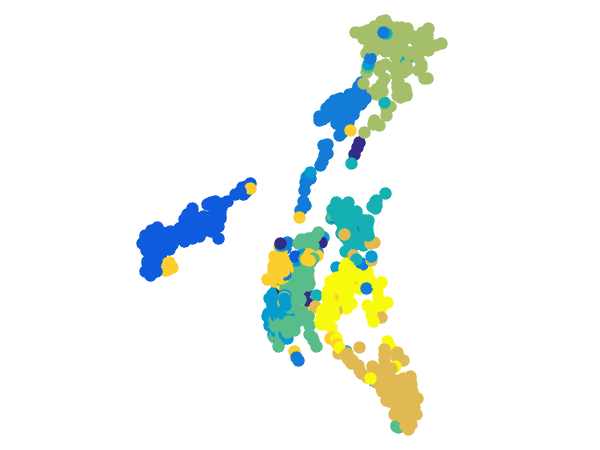}}
\centerline{\footnotesize{(b) Layer-1 output features}}
\end{center} 
\end{minipage}
\begin{minipage}[b]{0.19\linewidth}
\begin{center}
\centerline{\includegraphics[width=\columnwidth]{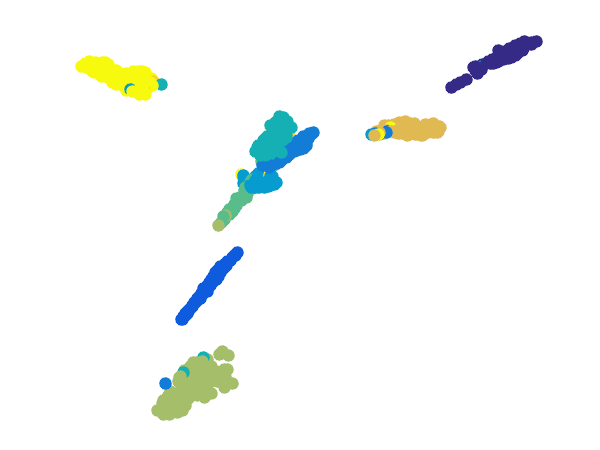}}
\centerline{\footnotesize{(c) Layer-2 output features}}
\end{center}
\end{minipage}
\begin{minipage}[b]{0.19\linewidth}
\begin{center}
\centerline{\includegraphics[width=\columnwidth]{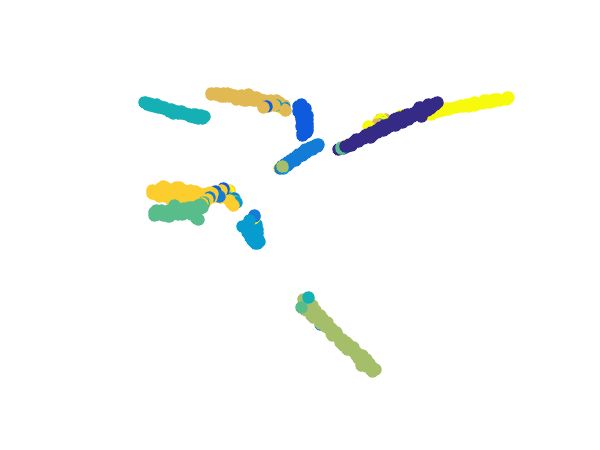}}
\centerline{\footnotesize{(d) Layer-4 output features}}
\end{center}
\end{minipage}
\begin{minipage}[b]{0.19\linewidth}
\begin{center}
\centerline{\includegraphics[width=\columnwidth]{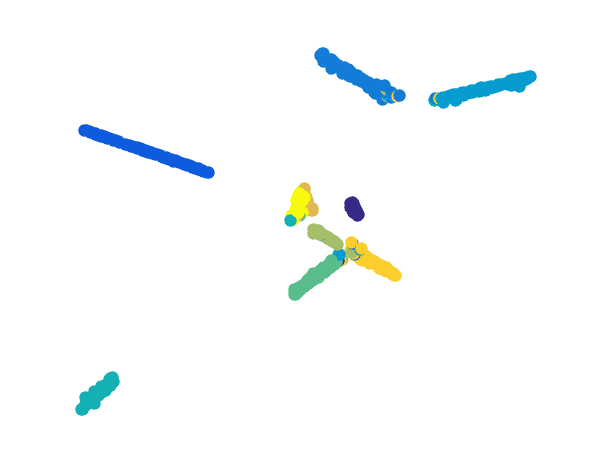}}
\centerline{\footnotesize{(e) Layer-8 output features}}
\end{center}
\end{minipage}
\vspace{-3mm}
\caption{\footnotesize{t-SNE visualization of different features on MNIST training samples, where different colors denote different classes. Clearly, for this network the output features with more hidden layers are better separated, \ie layer-8 output features (before rounding) are the best.
}}
\label{fig:mnist-stne}
\vspace{-2mm}
\end{figure*}


\subsection{Discussion}\label{ssec:mnist}
%
We analyze the behavior of VDSH training algorithm in Alg. \ref{alg:cd} with a small DNN of 8 hidden layers and 64 nodes (or neurons) per layer on the MNIST \cite{lecun-mnisthandwrittendigit-2010} dataset. For simplicity, all training parameters are set beforehand. Each subproblem in Alg. \ref{alg:cd} is optimized with subgradient descent.


\noindent
{\bf (\rom{1}) Empirical convergence:} Theoretically our VDSH is not guaranteed to converge to local minima. Nevertheless, empirically ADMM works well even if the objectives are nonconvex as observed in \cite{hong2014convergence}.  
Note that the Lagrangian dual variables for $\mathbf{z}$'s (\ie $\mathbb{E}_i(\beta\|\mathbf{u}_{i,m}\|_2),\,\forall m$) and $\boldsymbol{\theta}$'s (\ie $\mathbb{E}_i(\gamma\|\mathbf{v}_{i,m}\|_2), \forall m$) will converge when $\mathbf{z}_{i,m} = f(\mathbf{z}_{i,m-1};\boldsymbol{\theta}^{(m)})$ and $ \tilde{\boldsymbol{\theta}}_i^{(m)}=\boldsymbol{\theta}^{(m)},\,\forall i, \forall m,$ holds respectively. This motivates us to plot the mean of the $\ell_2$ norm of the Lagrangian dual variables to demonstrate the empirical convergence behavior of our VDSH.

Fig. \ref{fig:convergence} depicts the empirical convergence behavior for each hidden layer. Intuition suggests that small dual update steps (\eg $\beta=10^{-5}, 10^{-3}$) lead to slow convergence, which we see empirically in slow change in terms of mean value. On the other hand large steps (\eg $\beta=10$) can lead to zigzag behavior around a local optimum. For an appropriate step size (\eg $\beta=10^{-1}$), we do see 
smooth convergence at all the layers.


Interestingly, for all the four different dual steps, {\em all eight layers tend to show similar convergence rates.} For instance, in Fig. \ref{fig:convergence}(c) where $\beta=10^{-1}$, all curves tend to be relatively flat by iteration 100. 
This implies larger changes at front layers and small changes at final layers in our network, leading to faster convergence. This in turn implies that {\em our training algorithm for VDSH has the potential to overcome the vanishing gradient issue in backprop\footnote{For graphical comparison on convergence rate, please refer to \url{http://neuralnetworksanddeeplearning.com/chap5.html}}}. Similar behavior has been observed for $\boldsymbol{\theta}$.


We visualize the output features from different layers with $\beta=10^{-1}$ at 100 iterations using t-SNE \cite{van2008visualizing} in Fig.~\ref{fig:mnist-stne}. As the number of layers increases, the data evidently forms clearer clusters, indicating that our VDSH not only encodes data effectively but also converges at each layer. 

\begin{figure}[t]
\begin{minipage}[b]{0.49\linewidth}
 \begin{center}
 \centerline{\includegraphics[width=\columnwidth]{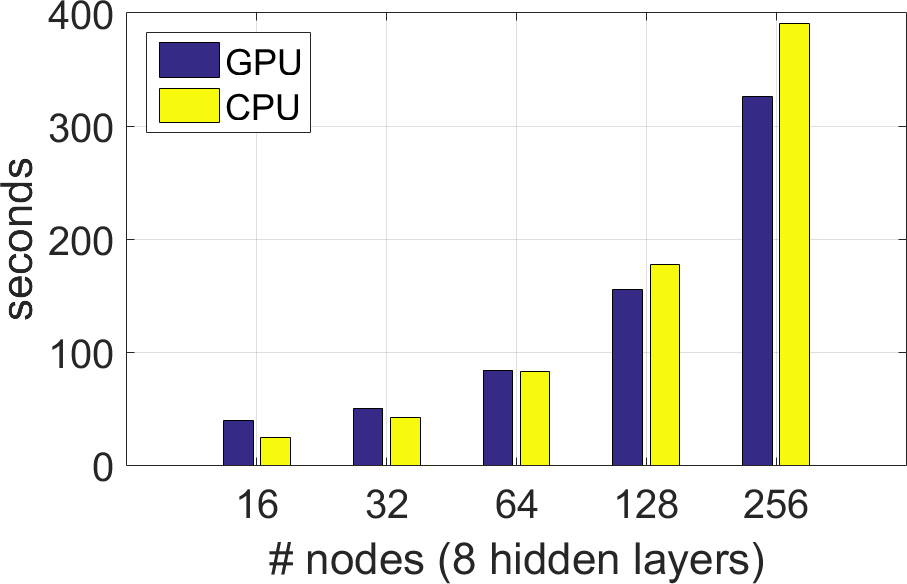}}
\centerline{\footnotesize{(a)}} 
 \end{center}
\end{minipage}
\begin{minipage}[b]{0.49\linewidth}
\begin{center}
\centerline{\includegraphics[width=\columnwidth]{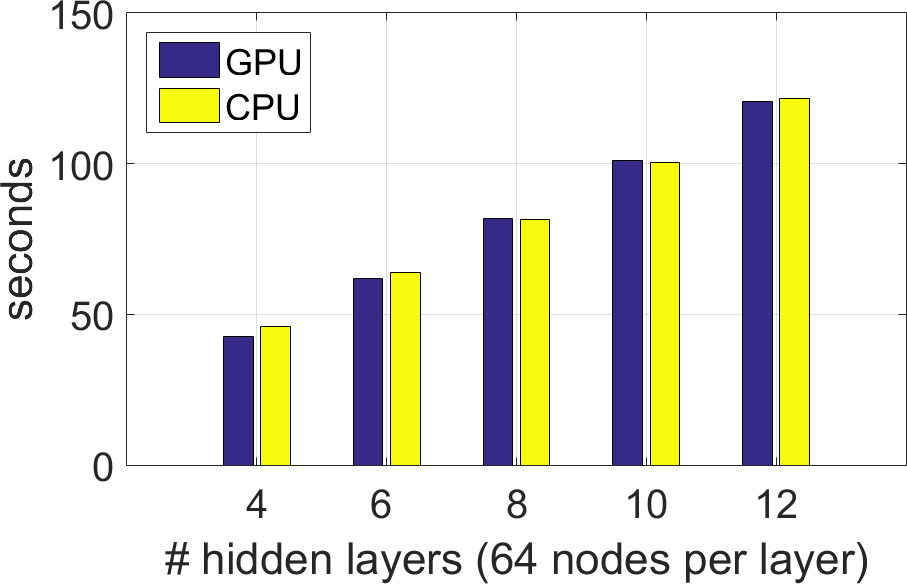}}
\centerline{\footnotesize{(b)}} 
\end{center} 
\end{minipage}
\vspace{-3mm}
\caption{\footnotesize{Actual training time comparison using CPU and GPU by {\bf (a)} training 8 hidden layer DNNs with different number of nodes per layer, and {\bf (b)} training DNNs with various hidden layers but 64 nodes per layer.
}}
\label{fig:time}
\vspace{-2mm}
\end{figure}

\begin{figure*}[t]
\begin{minipage}[b]{0.24\linewidth}
 \begin{center}
 \centerline{\includegraphics[width=.9\columnwidth]{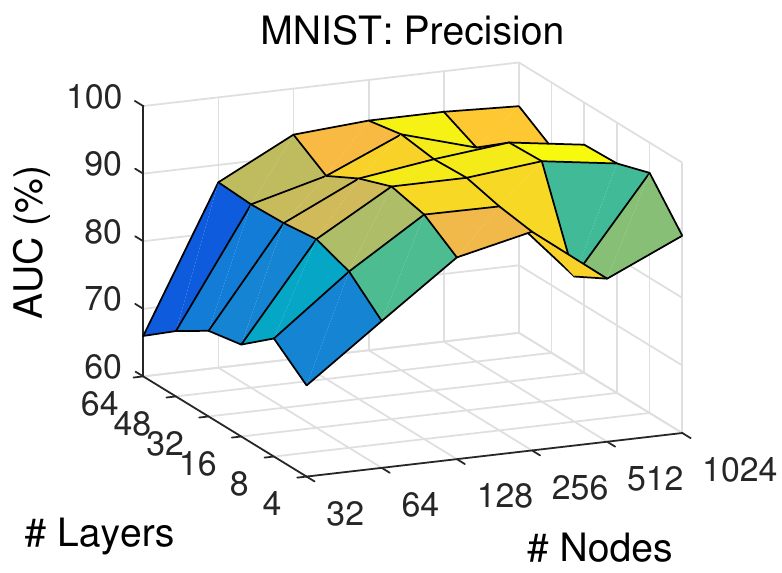}}
\centerline{\footnotesize{(a)}} 
 \end{center}
\end{minipage}
\begin{minipage}[b]{0.24\linewidth}
\begin{center}
\centerline{\includegraphics[width=.9\columnwidth]{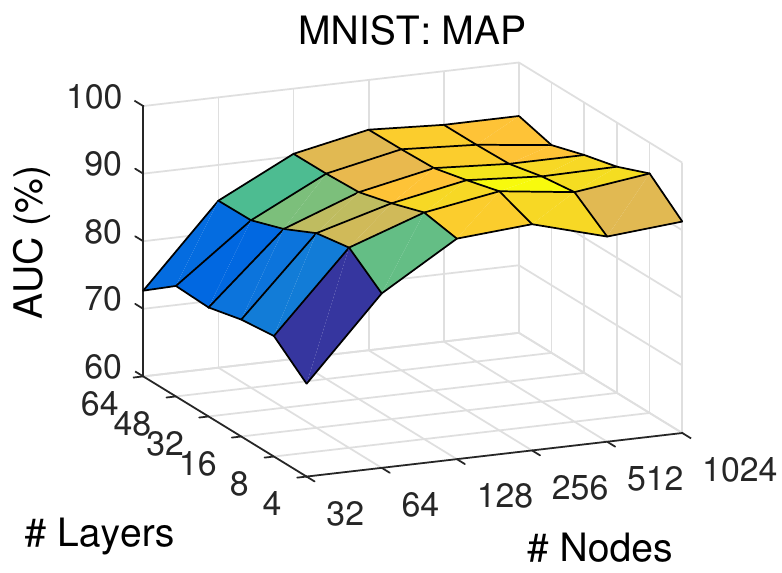}}
\centerline{\footnotesize{(b)}} 
\end{center} 
\end{minipage}
\begin{minipage}[b]{0.24\linewidth}
 \begin{center}
 \centerline{\includegraphics[width=.9\columnwidth]{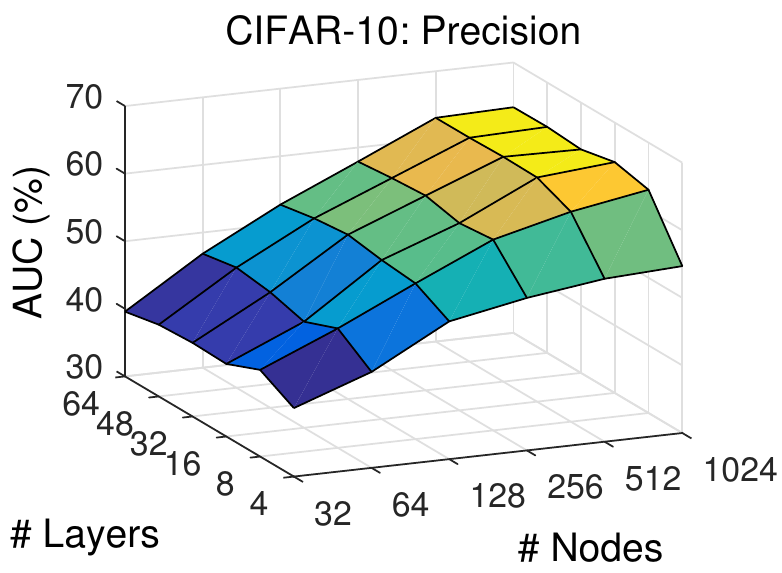}}
\centerline{\footnotesize{(c)}} 
 \end{center}
\end{minipage}
\begin{minipage}[b]{0.24\linewidth}
\begin{center}
\centerline{\includegraphics[width=.9\columnwidth]{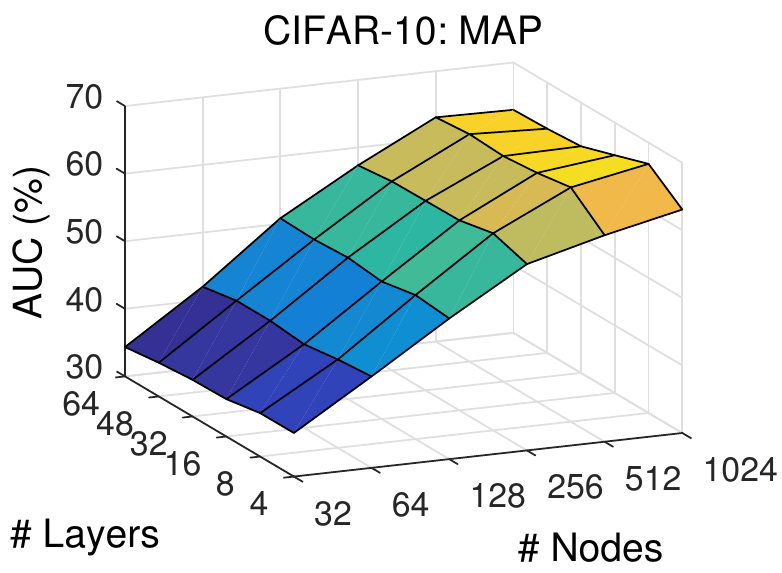}}
\centerline{\footnotesize{(d)}} 
\end{center} 
\end{minipage}
\vspace{-3mm}
\caption{\footnotesize{Network evaluation using default features (\ie pixel intensities and GIST) on MNIST and CIFAR-10. {\bf (a, c)} AUC of the precision \vs code-length curve \wrt varying number of layers and dimensions. {\bf (b, d)} AUC of the MAP \vs code-length curve \wrt varying number of layers and dimensions.
}}
\label{fig:NET_DIM_sensitivity}
\vspace{-3mm}
\end{figure*}

\noindent
{\bf (\rom{2}) Computational complexity:} 
The computational complexity of VDSH is $O(\sum_{m=0}^{M}D_mD_{m+1}N)$ where $D_0=d$ denotes the input dimension, $D_{M+1}=N_c$ denotes output dimension (\ie the number of classes), and $N$ the number of training samples. This follows from the fact that the computational complexity of training VDSH is proportional to training each individual two-layer substructure (see Fig. \ref{fig:targetprop}) on account of our ADMM-style decomposition. Now since information goes through the substructure back and forth with subgradient descent updates, 
the computational complexity of a substructure per data sample corresponding to layers $m,\,m+1$ 
grows as $O(D_mD_{m+1})$.


We depict the speed of training using un-optimized MATLAB implementation\footnote{Our code can be downloaded at \url{https://zimingzhang.wordpress.com/}.} in Fig. \ref{fig:time}. All training parameters are set as default. The CPU and GPU used for comparison are i7-4930MX@3GHz and Quadro K2100M, respectively. The timing behavior using either CPU or GPU in both plots supports our computational complexity analysis above: in (a) the timing is roughly quadratic in the number of nodes, and in (b) the timing is roughly linear in the number of hidden layers.

We also compare our method with backprop in terms of computational time. To train a shallow model with 4 hidden layers and 64 nodes per layer, our training speed is about 20 times faster than backprop while achieving similar performance. However, to train a deeper model with 48 hidden layers and 256 nodes per layer, our training algorithm converges within 1 hour, while backprop has not converged within weeks.

\section{Experiments}\label{sec:exp}
In this section, we compare our VDSH with state-of-the-art supervised hashing methods, including SDH \cite{shen2015supervised}, BRE \cite{kulis2009learning}, MLH \cite{norouzi2011minimal}, CCA-ITQ \cite{gong2013iterative}, KSH \cite{liu2012supervised}, FastHash \cite{lin2014fast}, DSRH \cite{zhao2015deep}, DSCH \cite{DBLP:journals/corr/ZhangLZZZ15} and DRSCH \cite{DBLP:journals/corr/ZhangLZZZ15} on image retrieval tasks. Following the evaluation protocols used in previous supervised hashing methods (e.g. \cite{shen2015supervised, DBLP:journals/corr/ZhangLZZZ15}), each dataset is split into a large retrieval database and a small query set. The entire retrieval database is used to train the hashing models unless otherwise specified. The lengths of output hash codes vary from 16 to 128 bits. The retrieval performance on the query set is evaluated using mean average precision (MAP) and precision (or recall) within Hamming radius 2. All the data samples are normalized to have unit length. For simplicity, our networks all have the same number of nodes in each hidden layer. We tune our network architectures as well as training parameters using cross validation on training data, and report our performance on the query data using the best networks. Our experiments are all run on an Xeon E5-2696 v2 and a single GTX TITAN with un-optimized MATLAB implementation.

\subsection{Datasets and Setup}
We test VDSH mainly on three benchmark datasets for image retrieval tasks with learned hash functions: MNIST, CIFAR-10 \cite{krizhevsky2009learning}, and NUS-WIDE \cite{nus-wide-civr09}. Our method learns the mapping function from image features to hash codes, equivalent to learning from image pixels implicitly by composition of functions.

MNIST contains 70K gray-scale handwritten digit images with $28\times28$ pixels from ``0'' to ``9''. Following \cite{shen2015supervised}, we randomly sample 100 images per class to form a 1K image query set, and use the rest 69K images as the training and retrieval database. By default each image is represented by a 784-dim vector consisting of its pixel intensities.

CIFAR-10 contains 60K color images of resolution of $32\times32$ pixels from 10 object classes, with 6K images per class. Following \cite{shen2015supervised}, we randomly sample 100 images per class as the query set and use the rest 59K images as the training and retrieval set. As default features, each image is presented by a 512-dim GIST \cite{oliva2001modeling} feature vector.

NUS-WIDE contains about 270K images collected from the web. It is a {\em multi-label} dataset where each image is associated with one or more of the 81 semantic concepts. Each image is represented by a 500-dim bag-of-words feature vector that is provided in the dataset. Following \cite{shen2015supervised}, we only consider the 21 most frequent concept labels and randomly sample 100 images per label to form the query set. The remaining images are used as the training and retrieval set. Two images are considered as a true match if they share at least one common label. 

\begin{figure*}
\begin{minipage}[b]{0.23\linewidth}
 \begin{center}
 \centerline{\includegraphics[width=.9\columnwidth]{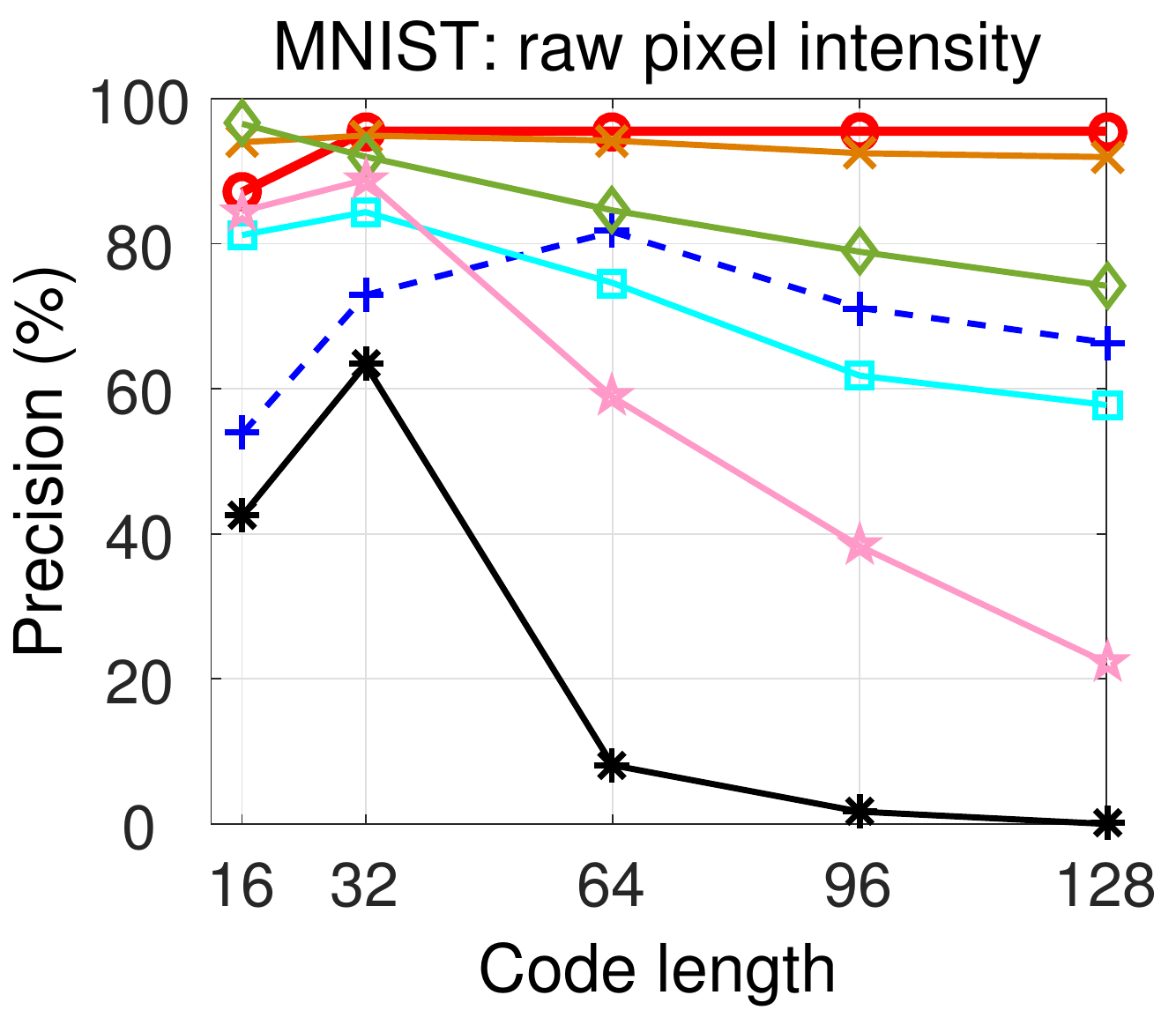}}
\centerline{\footnotesize{(a)}} 
 \end{center}
 \vspace*{-5mm}
\end{minipage}
\begin{minipage}[b]{0.23\linewidth}
\begin{center}
\centerline{\includegraphics[width=.9\columnwidth]{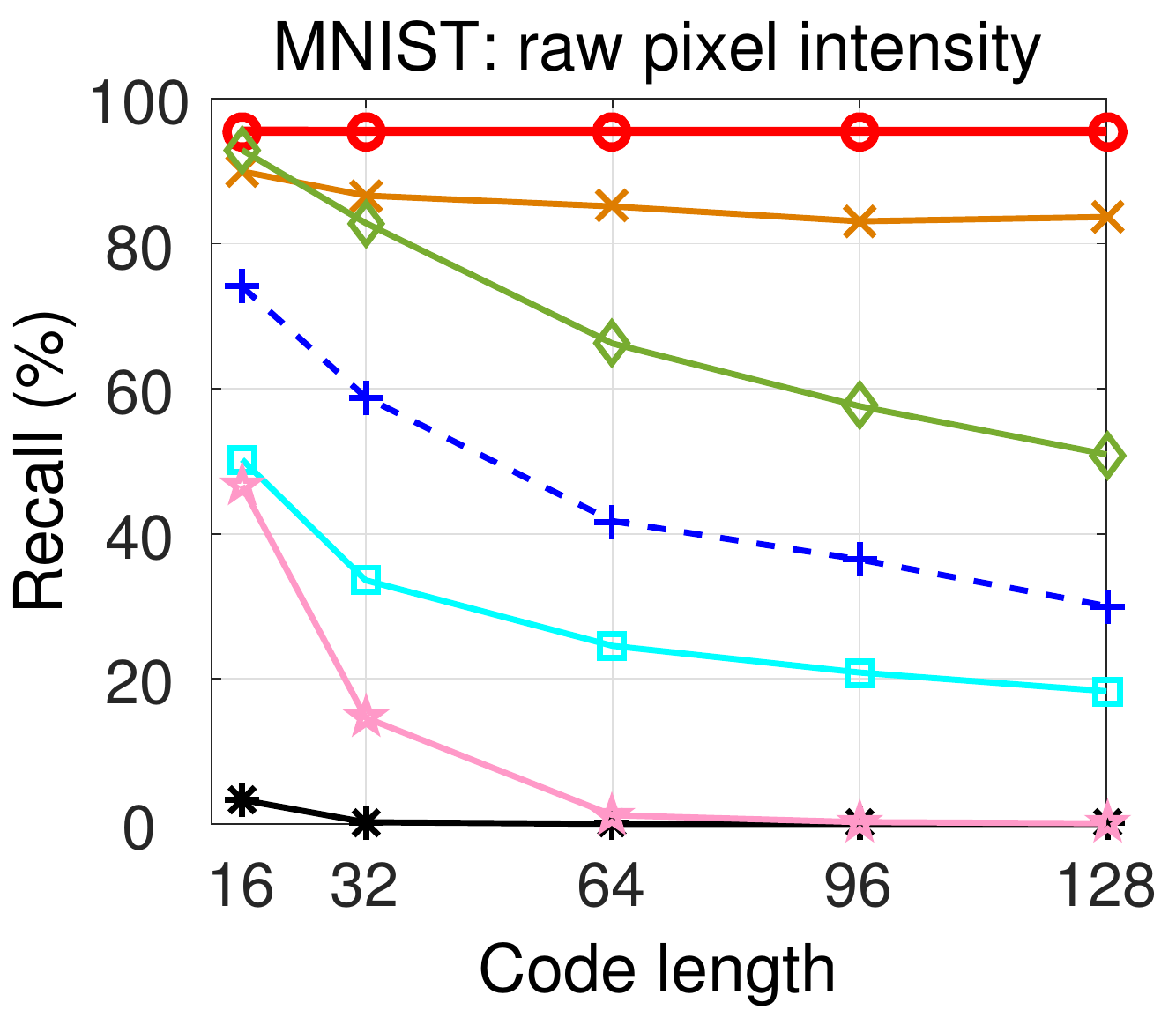}}
\centerline{\footnotesize{(b)}} 
\end{center} 
 \vspace*{-5mm}
\end{minipage}
\begin{minipage}[b]{0.23\linewidth}
 \begin{center}
 \centerline{\includegraphics[width=.9\columnwidth]{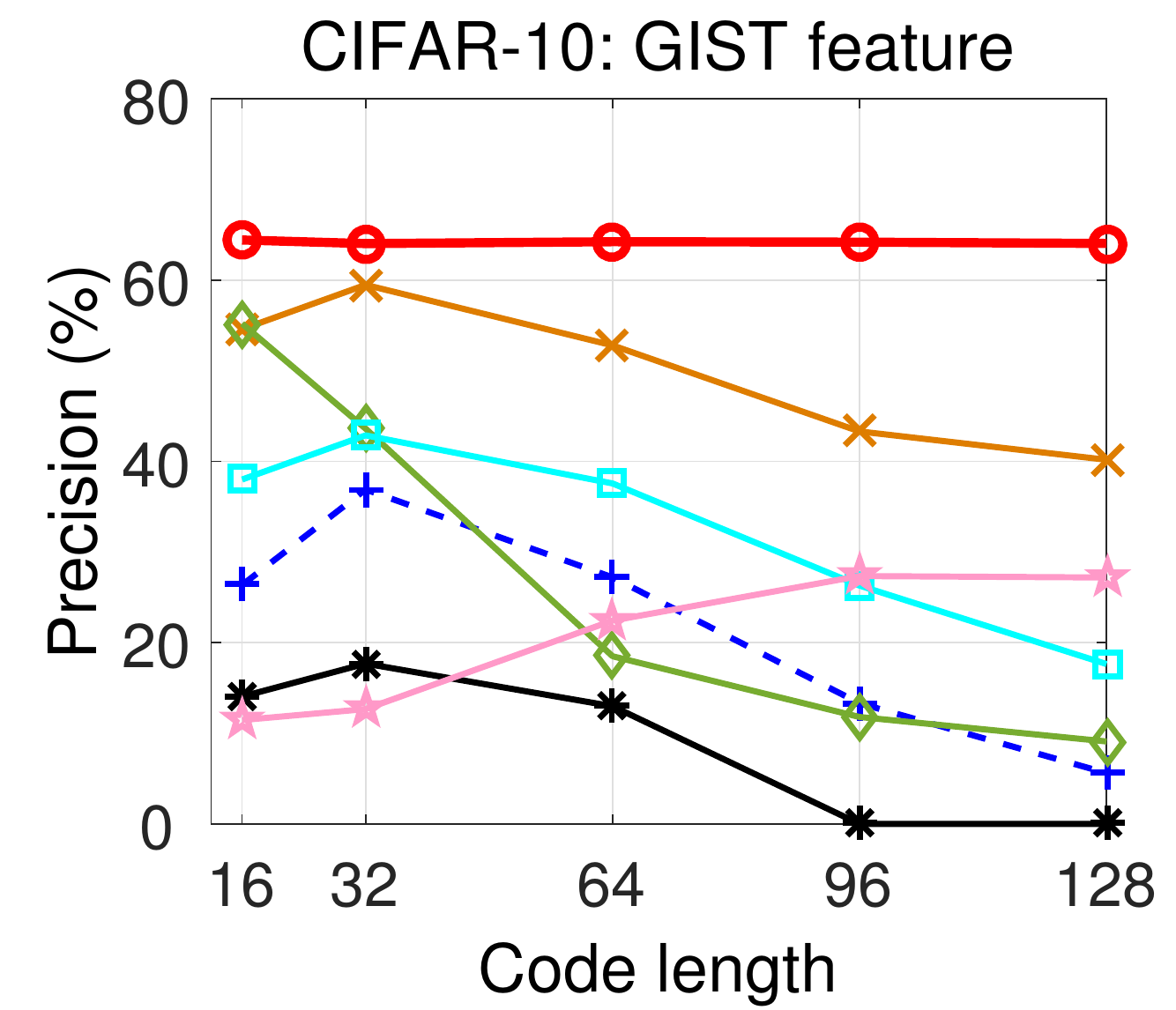}}
\centerline{\footnotesize{(c)}} 
 \end{center}
 \vspace*{-5mm}
\end{minipage}
\begin{minipage}[b]{0.30\linewidth}
\begin{center}
\centerline{\includegraphics[width=.9\columnwidth]{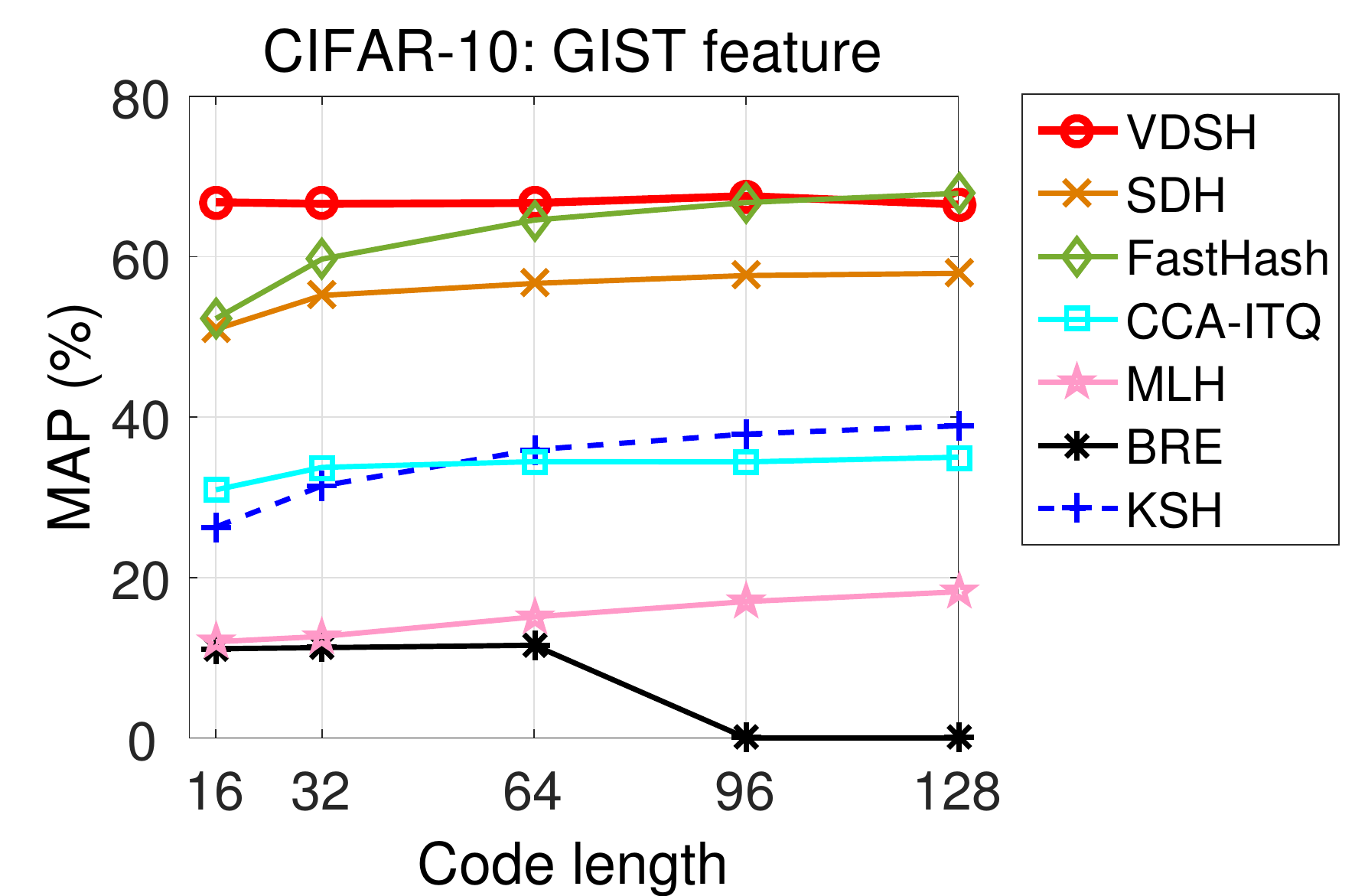}}
\centerline{\footnotesize{(d)}} 
\end{center} 
 \vspace*{-5mm}
\end{minipage}
\begin{minipage}[b]{0.23\linewidth}
 \begin{center}
 \centerline{\includegraphics[width=.9\columnwidth]{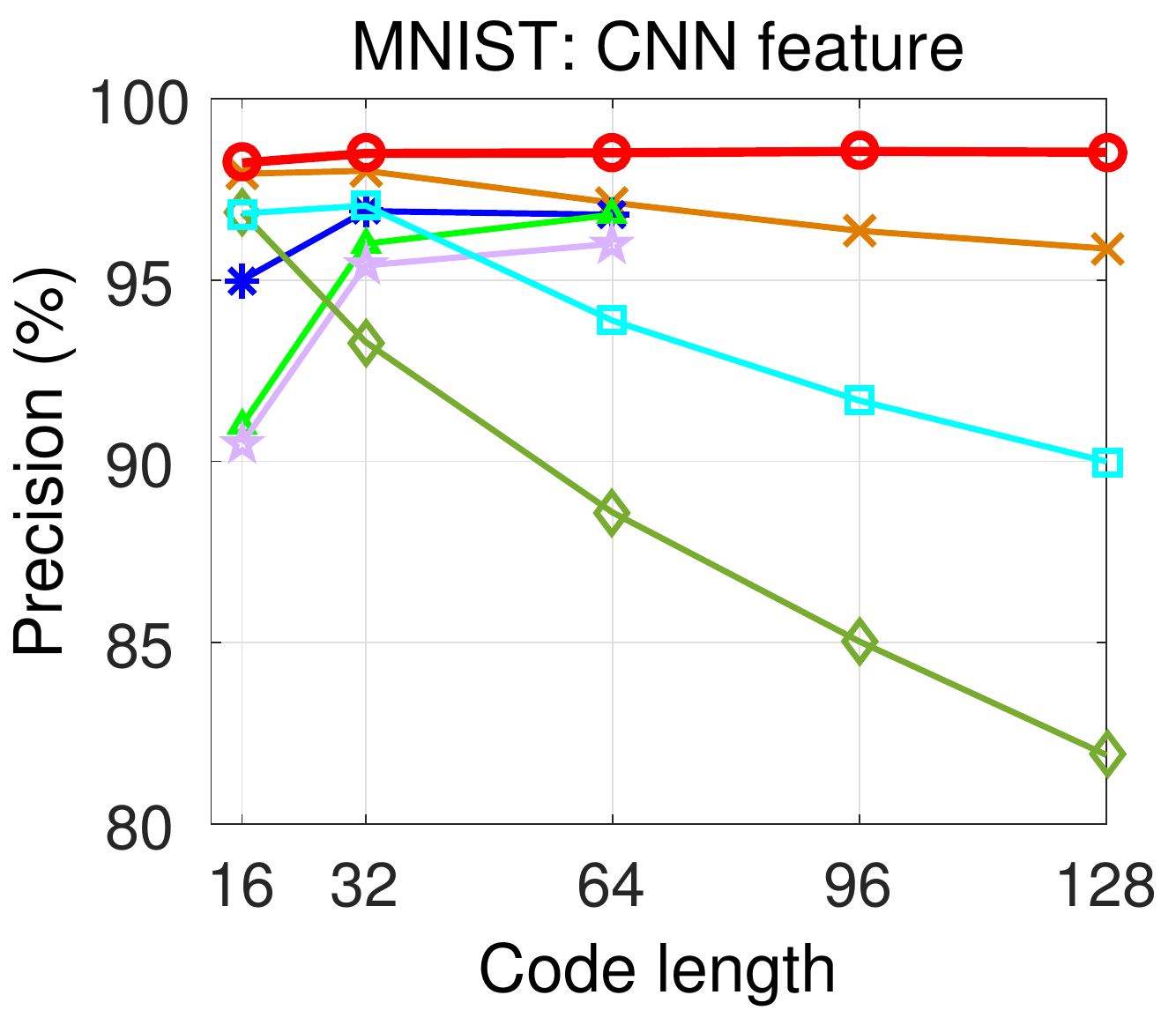}}
\centerline{\footnotesize{(e)}} 
 \end{center}
\end{minipage}
\begin{minipage}[b]{0.23\linewidth}
\begin{center}
\centerline{\includegraphics[width=.9\columnwidth]{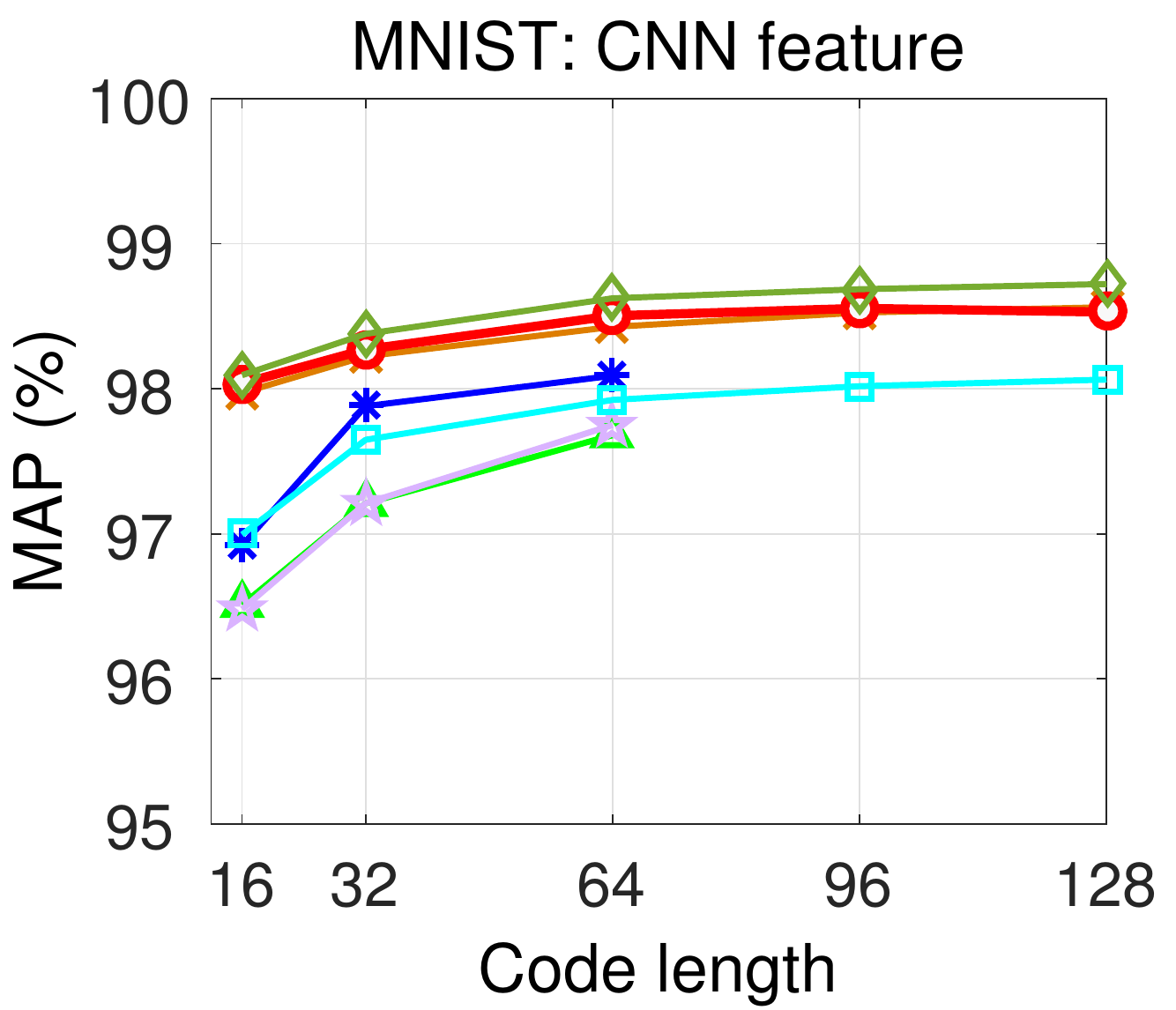}}
\centerline{\footnotesize{(f)}} 
\end{center} 
\end{minipage}
\begin{minipage}[b]{0.23\linewidth}
 \begin{center}
 \centerline{\includegraphics[width=.9\columnwidth]{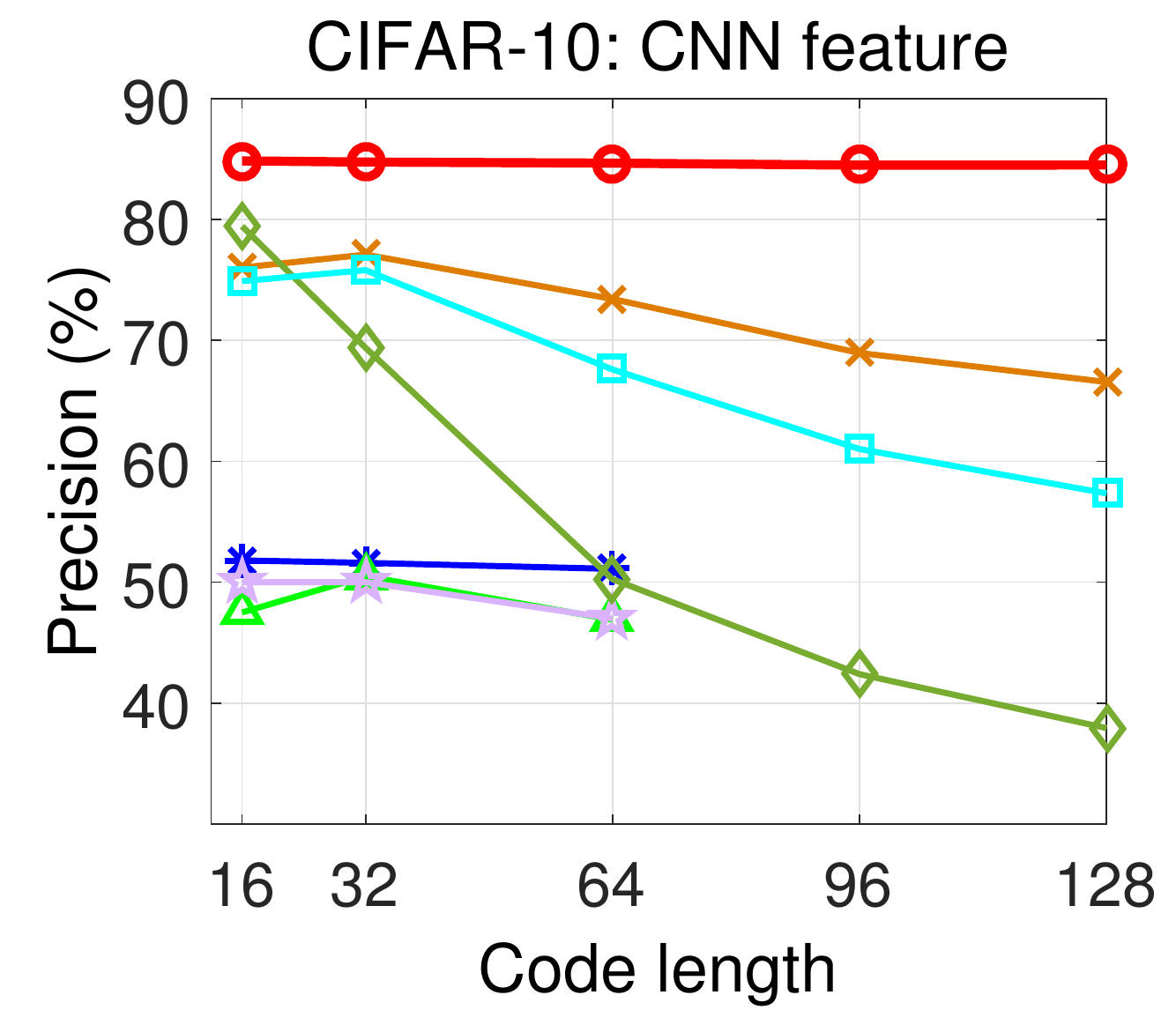}}
\centerline{\footnotesize{(g)}} 
 \end{center}
\end{minipage}
\begin{minipage}[b]{0.30\linewidth}
\begin{center}
\centerline{\includegraphics[width=.9\columnwidth]{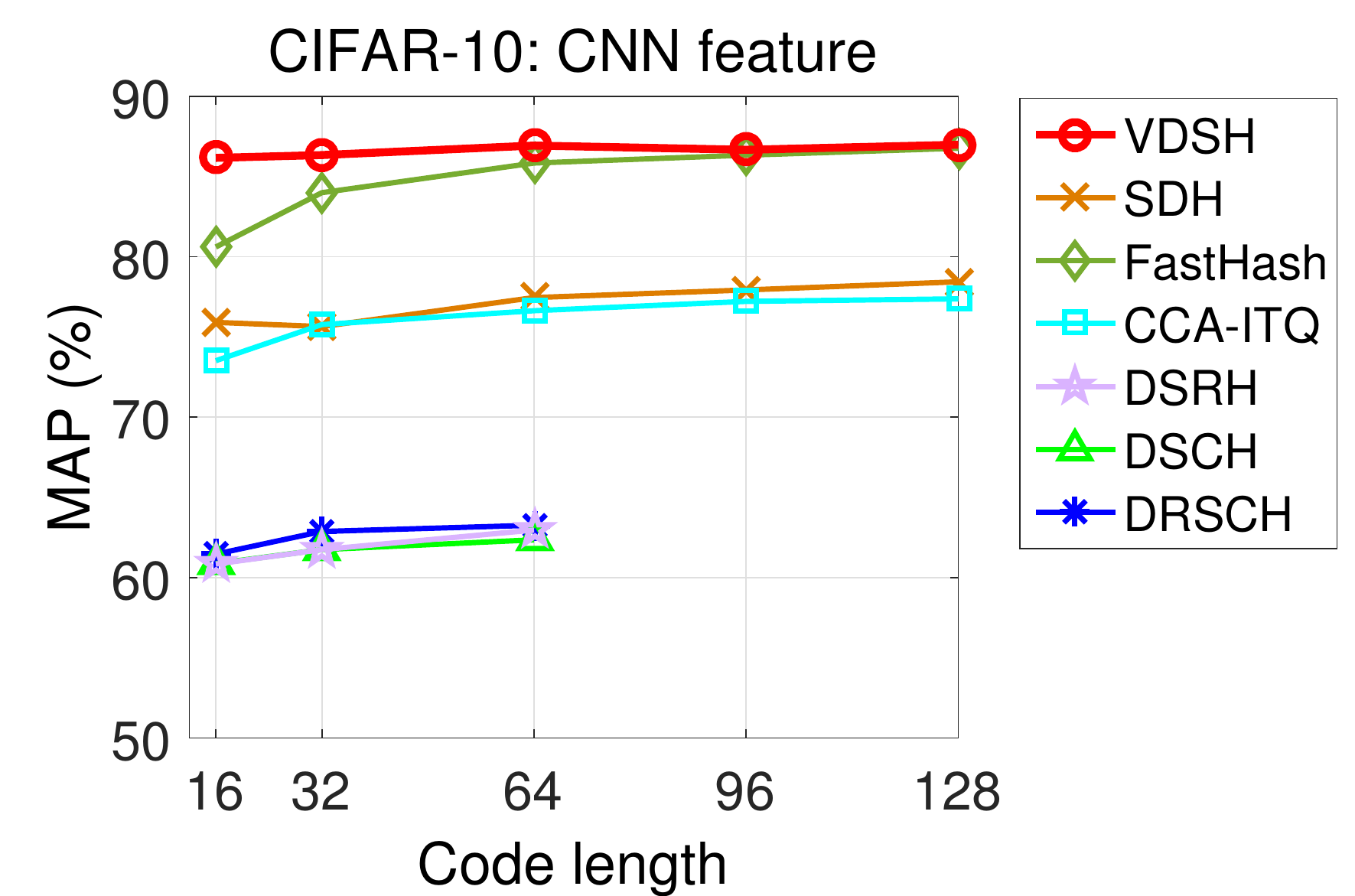}}
\centerline{\footnotesize{(h)}} 
\end{center} 
\end{minipage}
\vspace{-7mm}
\caption{Retrieval performance comparison on MNIST and CIFAR-10 within Hamming radius 2.} 		
\label{fig:PreRecMAP_mnist_cifar}
\vspace{-3mm}
\end{figure*}

\begin{figure*}[t]
\begin{minipage}[b]{0.24\linewidth}
 \begin{center}
 \centerline{\includegraphics[width=.9\columnwidth]{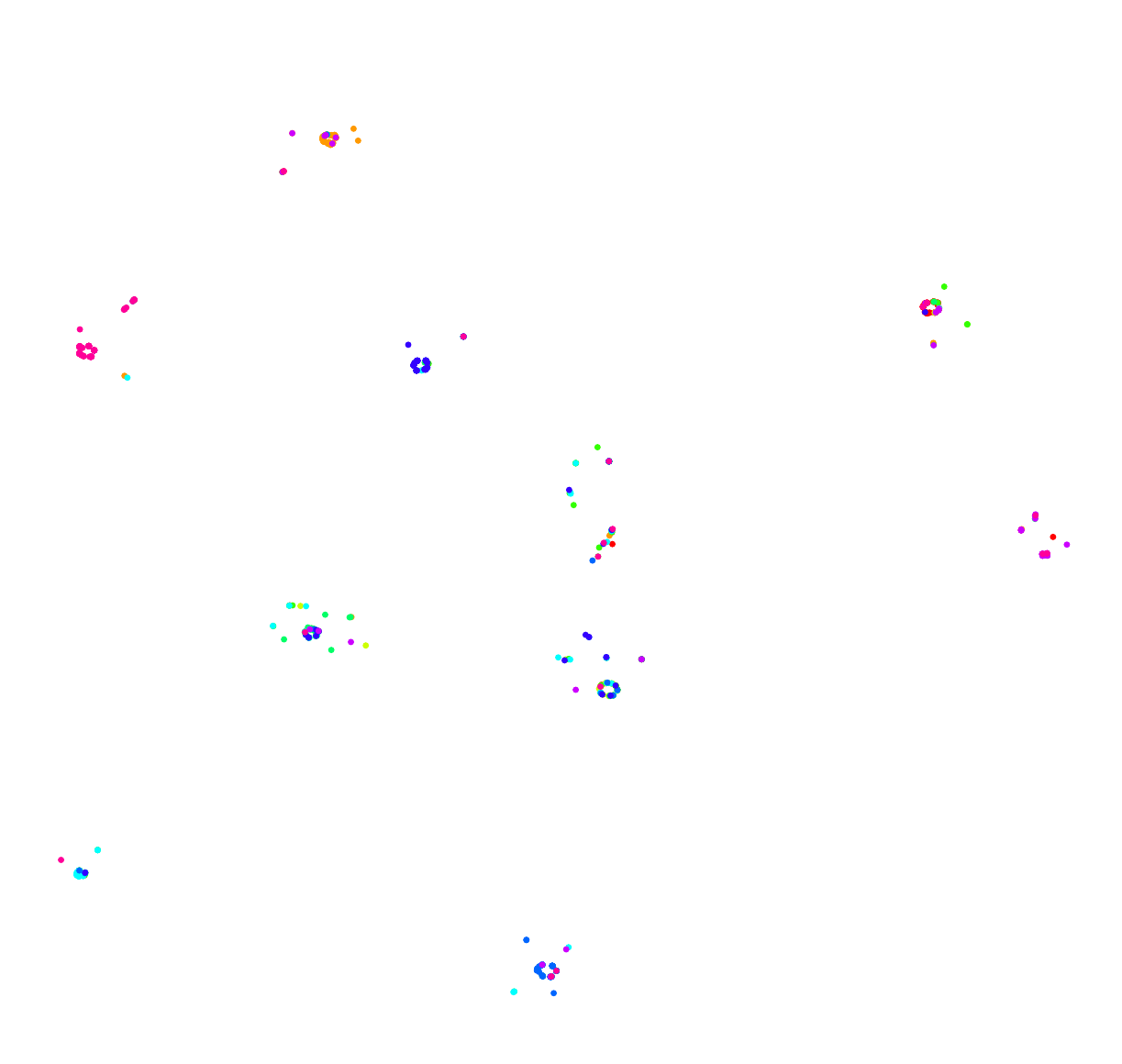}}
\centerline{\footnotesize{(a) VDSH: GIST feature}} 
 \end{center}
\end{minipage}
\begin{minipage}[b]{0.24\linewidth}
\begin{center}
\centerline{\includegraphics[width=.9\columnwidth]{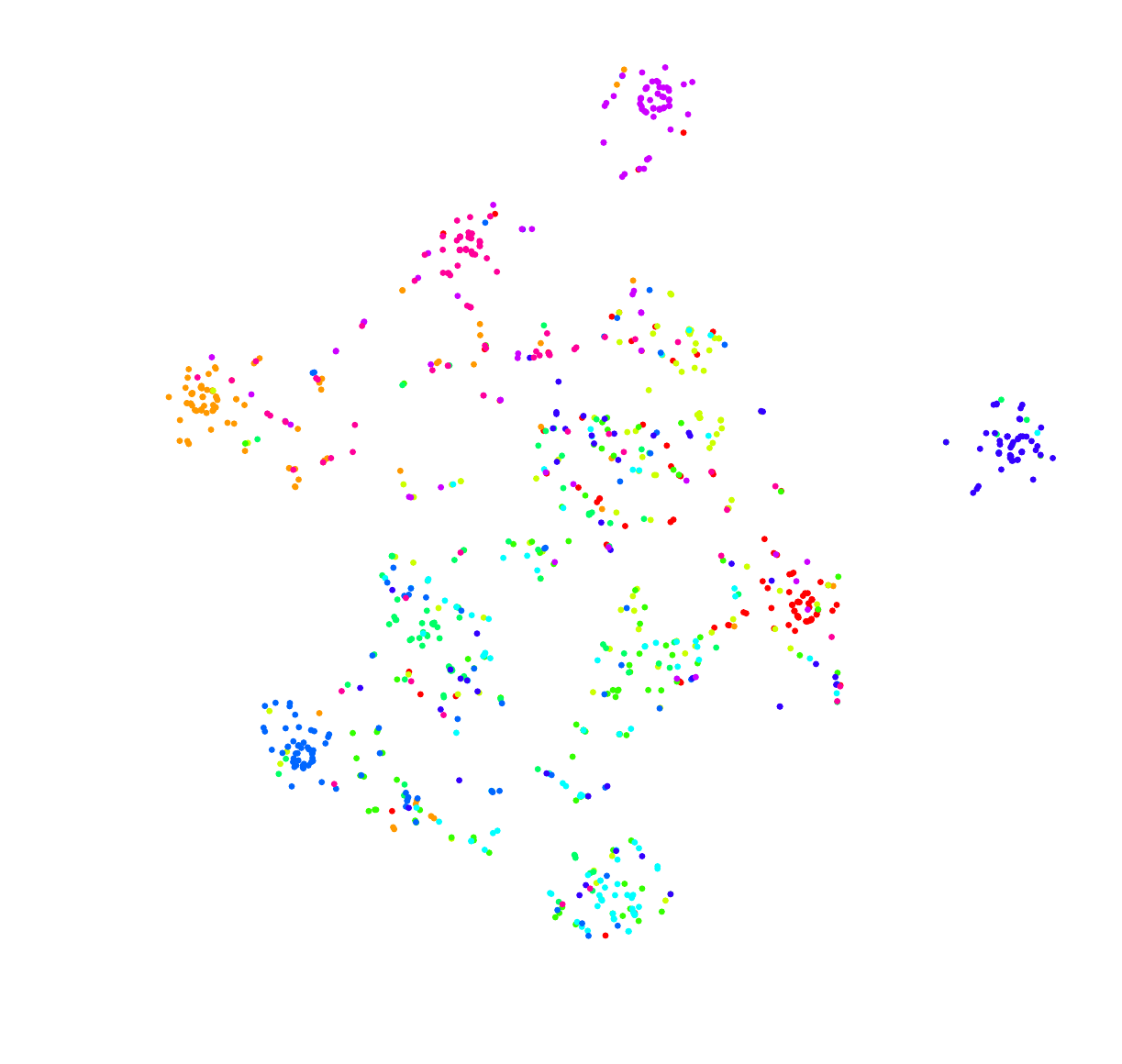}}
\centerline{\footnotesize{(b) SDH: GIST feature}} 
\end{center} 
\end{minipage}
\begin{minipage}[b]{0.24\linewidth}
 \begin{center}
 \centerline{\includegraphics[width=.9\columnwidth]{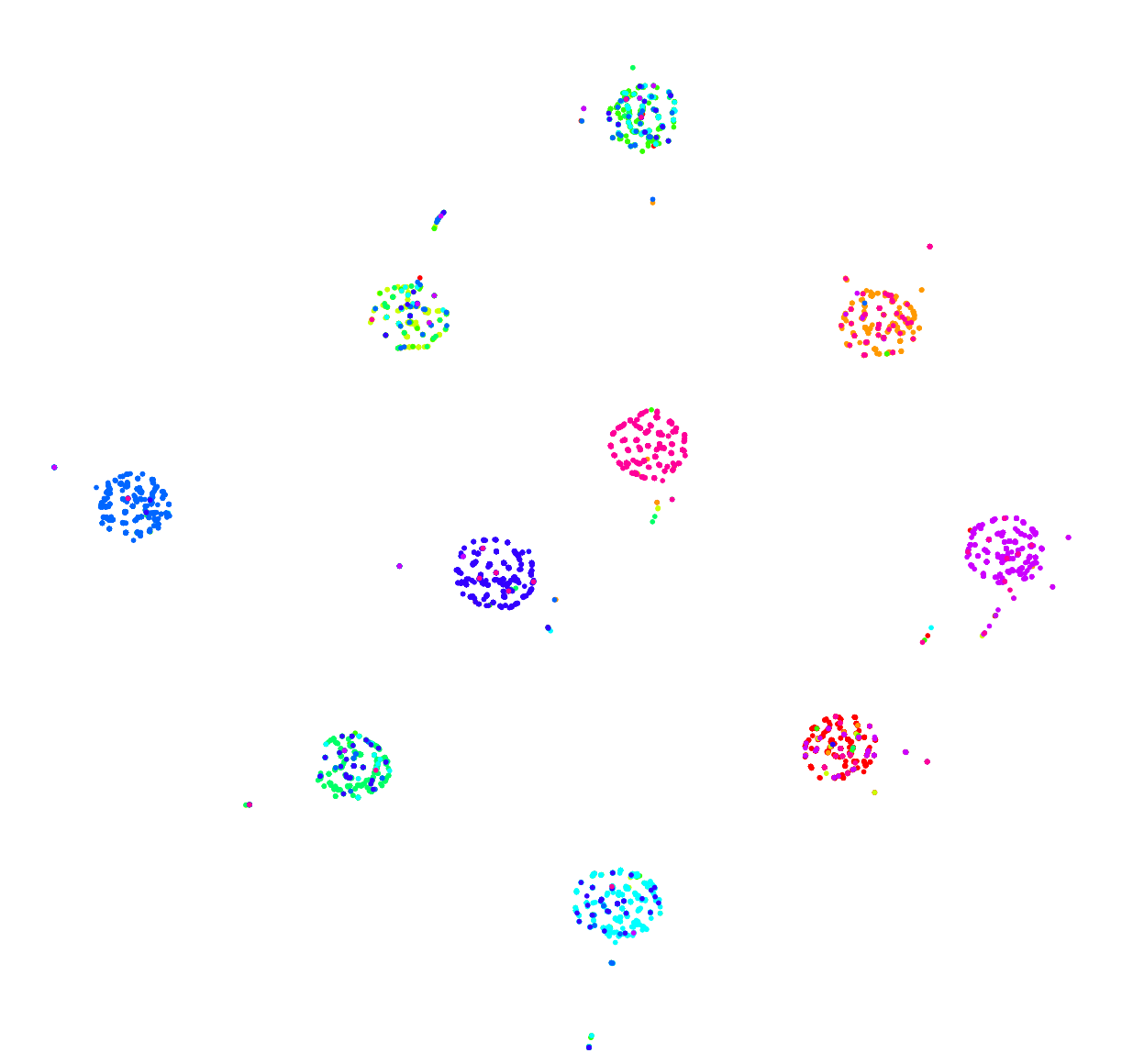}}
\centerline{\footnotesize{(c) VDSH: CNN feature}} 
 \end{center}
\end{minipage}
\begin{minipage}[b]{0.24\linewidth}
\begin{center}
\centerline{\includegraphics[width=.9\columnwidth]{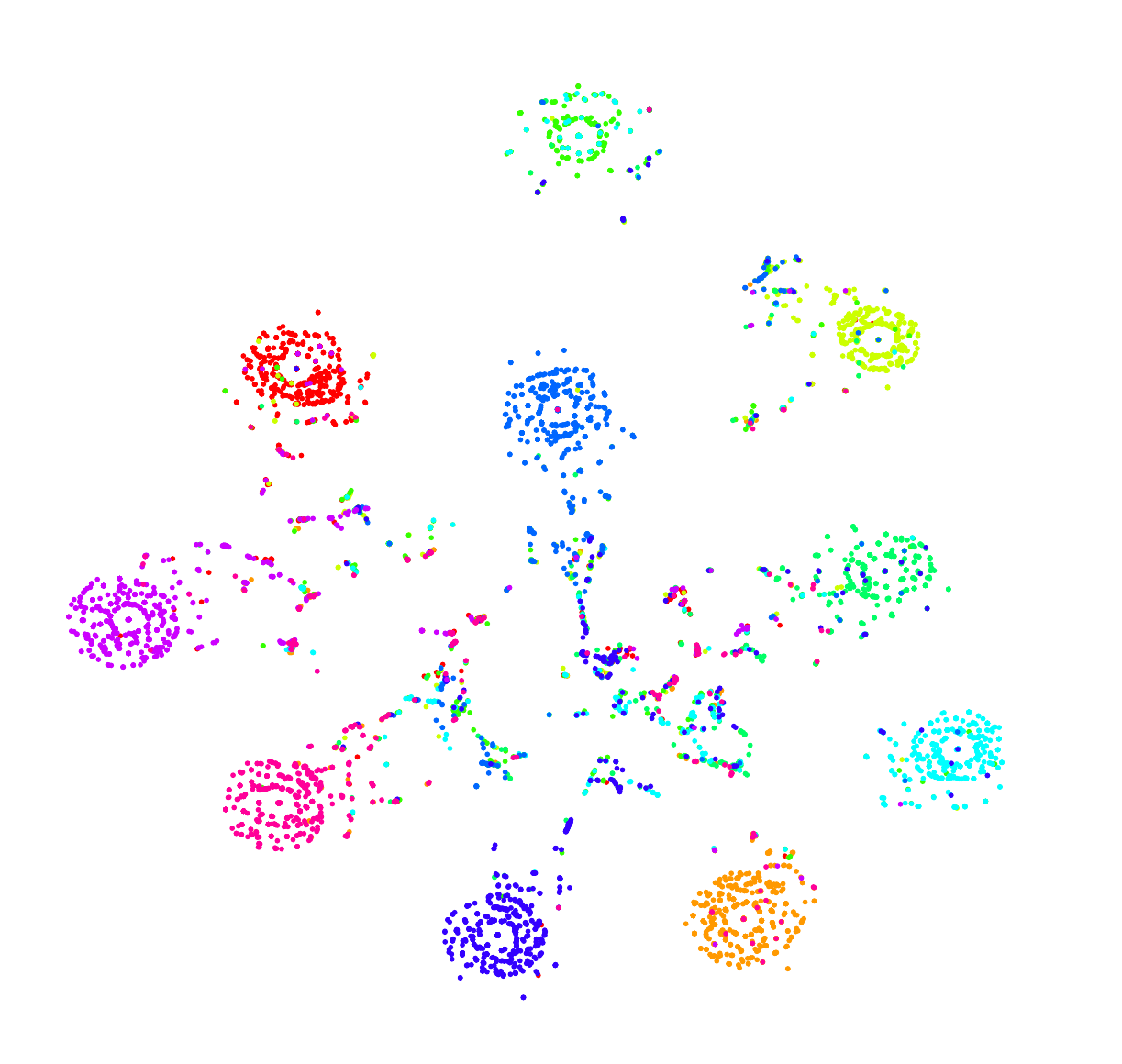}}
\centerline{\footnotesize{(d) SDH: CNN feature}} 
\end{center} 
\end{minipage}
\vspace{-1mm}
\caption{\footnotesize{t-SNE visualization of the 64-bit binary hash codes of all test images in CIFAR-10. {\bf (a-b)} or {\bf (c-d)} are plotted using the same images and scales.
}}
\label{fig:Visualization-CIFAR-10}
\vspace{-3mm}
\end{figure*}

\begin{figure*}[t]
\begin{minipage}[b]{0.23\linewidth}
 \begin{center}
 \centerline{\includegraphics[width=.9\columnwidth]{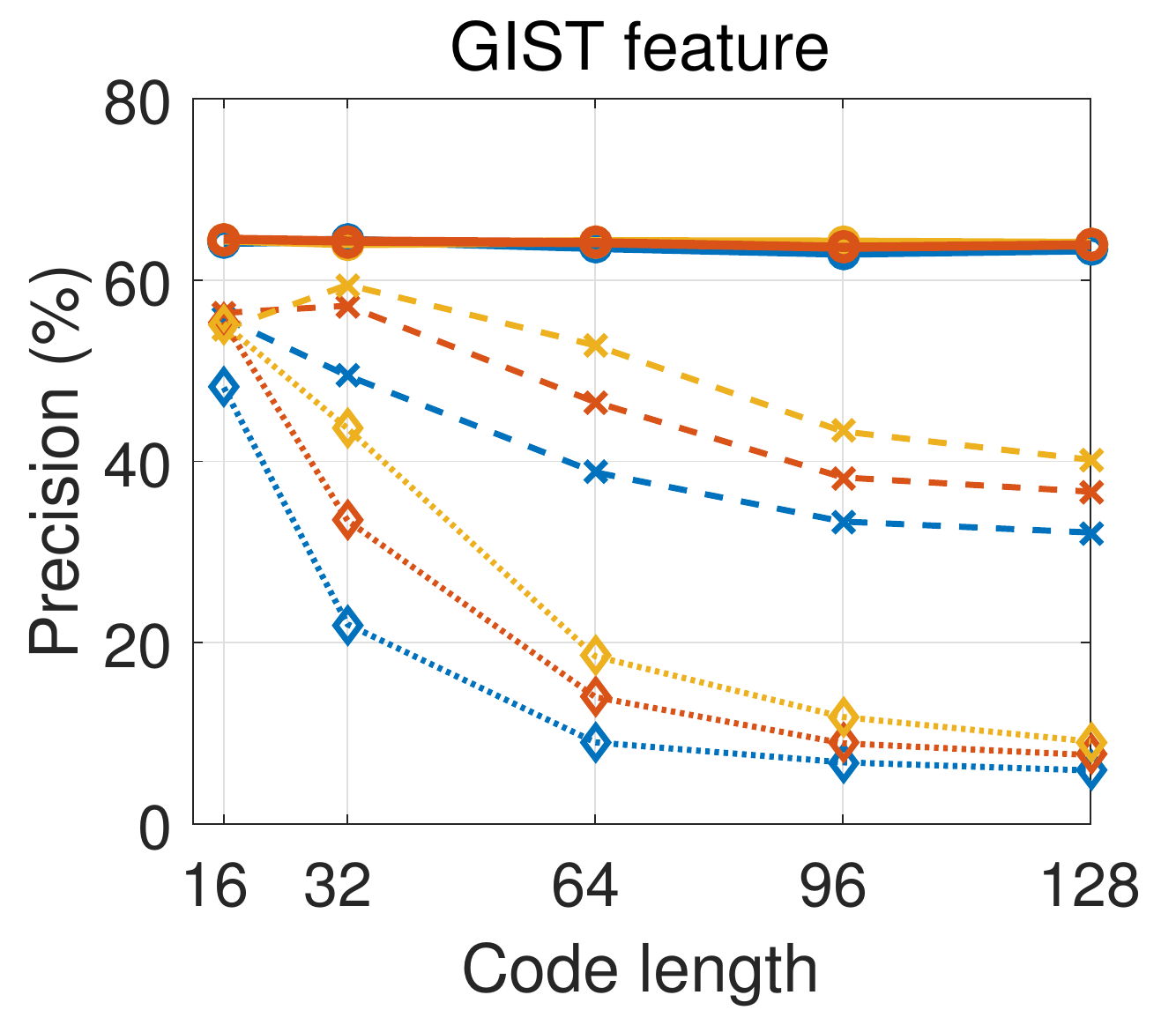}}
\centerline{\footnotesize{(a)}} 
 \end{center}
\end{minipage}
\begin{minipage}[b]{0.23\linewidth}
\begin{center}
\centerline{\includegraphics[width=.9\columnwidth]{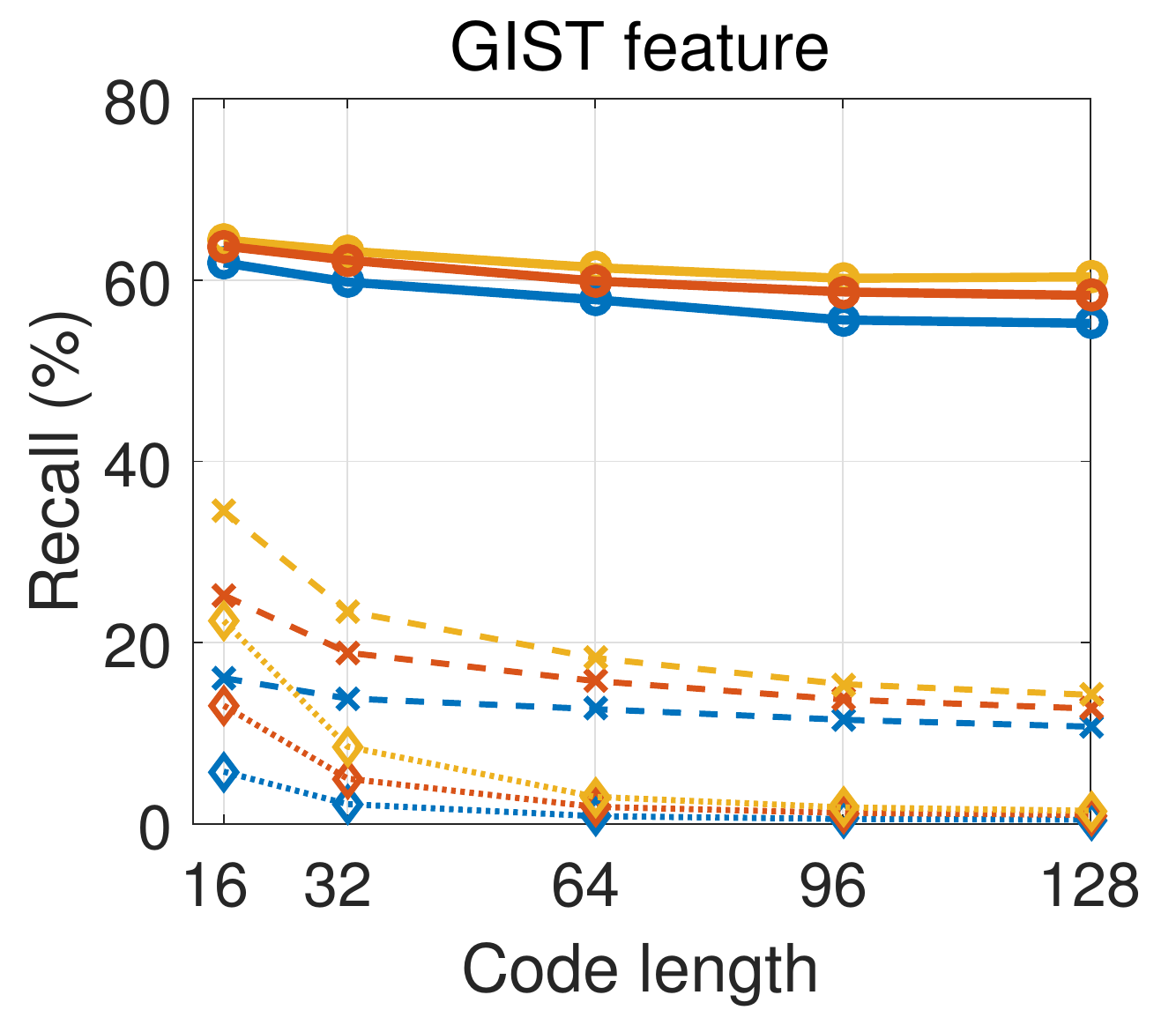}}
\centerline{\footnotesize{(b)}} 
\end{center} 
\end{minipage}
\begin{minipage}[b]{0.23\linewidth}
 \begin{center}
 \centerline{\includegraphics[width=.9\columnwidth]{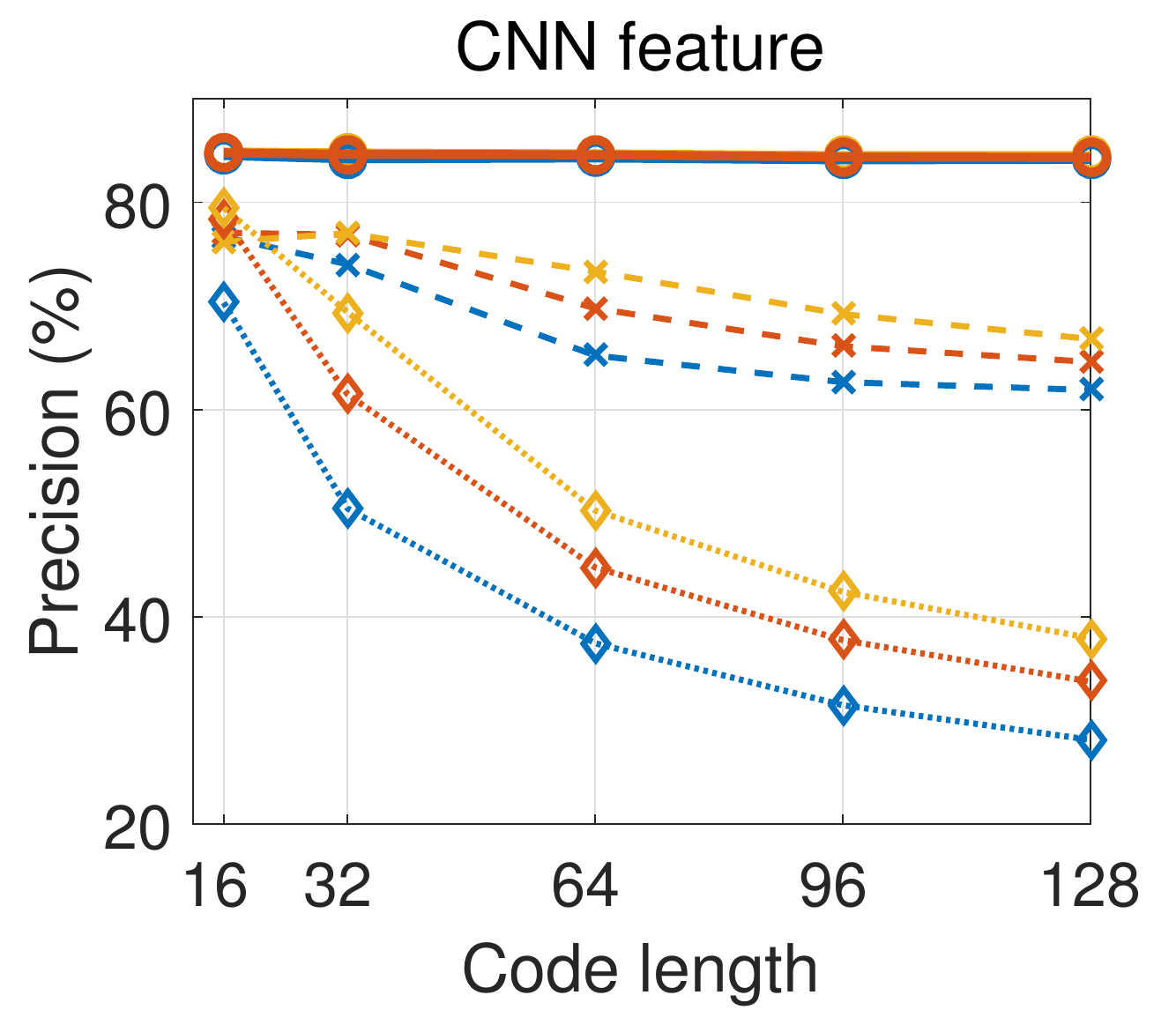}}
\centerline{\footnotesize{(c)}} 
 \end{center}
\end{minipage}
\begin{minipage}[b]{0.30\linewidth}
\begin{center}
\centerline{\includegraphics[width=.9\columnwidth]{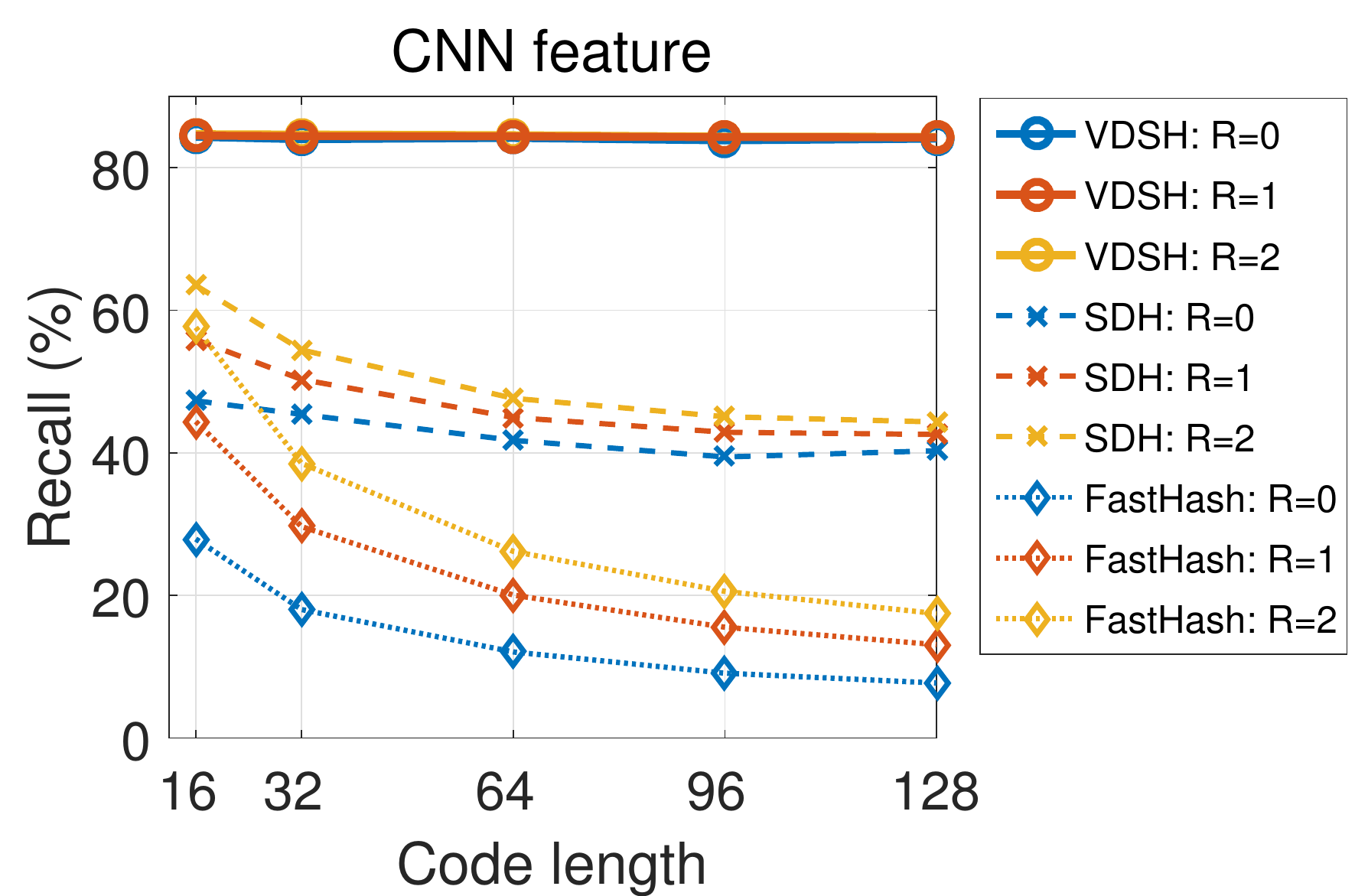}}
\centerline{\footnotesize{(d)}} 
\end{center} 
\end{minipage}
\vspace{-6mm}
\caption{\footnotesize{Precision-recall comparison on CIFAR-10 by varying Hamming radius (denoted by ``R'') using {\bf (a-b)} GIST features and {\bf (c-d)} CNN features.
}}
\label{fig:hamming}
\vspace{-3mm}
\end{figure*}

\subsection{Network Evaluation}

To explore the effect of different network architectures on the retrieval performance, we train a series of networks with varying depth from 4 to 64 hidden layers and dimension from 32 to 1024 nodes per layer, and report the Area-Under-Curve (\%) of the precision and MAP for varying code lengths in Fig. \ref{fig:NET_DIM_sensitivity} for MNIST and CIFAR-10. 
Note that for both metrics the plots on both datasets behave similarly, but the best networks for each dataset is different.

%
In general larger networks with more hidden layers and nodes per layer lead to better hash codes and better performance. 
The performance appears to saturate beyond a certain network size which in turn demonstrates the utility of regularization in preventing overly complicated models. In addition we also see that as the number of nodes/layers increases we obtain better retrieval performance. Intuitively, this makes sense because these numbers control the amount of information passing from one layer to the other.

\subsection{Performance Comparisons}
We compare VDSH with other supervised hashing methods in detail on MNIST and CIFAR-10, respectively. As our final models, we train a network with 48 hidden layers and 256 nodes per layer on MNIST, and a network with 16 hidden layers and 1024 nodes per layer on CIFAR-10. 
The training time for MNIST is about 15 minutes, and 6.6 milliseconds per sample for testing including hash code generation to retrieve a 69K-sample database. CIFAR-10 takes around 1 hour for training, and 4 milliseconds per sample to retrieve a 59K-sample database.

The comparison with default features is shown in Fig. \ref{fig:PreRecMAP_mnist_cifar} (a-d). 
Note that we are unable to use the full training set for BRE and KSH due to their huge memory requirements, and hence a 5K image subset is randomly sampled for these methods. We can see clearly that our VDSH significantly outputs the competitors by large margins. Also VDSH is more robust than others by maintaining very stable performance across increasing code lengths. 

In order to compare VDSH fairly with other deep hashing methods which learn the CNN features jointly with the hash codes, we utilize the pre-trained ``vgg-f'' model \cite{Chatfield14} to extract CNN features on MNIST and CIFAR-10 directly without any fine-tuning. We then apply VDSH, SDH, CCA-ITQ and FastHash on these CNN features to generate hash codes. Compared to fully optimized deep hashing methods such as DRSCH \cite{DBLP:journals/corr/ZhangLZZZ15}, this two-stage scheme has not been optimized for retrieval. The pre-learned CNN is agnostic to the hash codes that are intended to be generated.
%
We report the precision and MAP comparison in Fig. \ref{fig:PreRecMAP_mnist_cifar}(e-h) with the same experimental settings as in \cite{DBLP:journals/corr/ZhangLZZZ15} and \cite{zhao2015deep} for the CNN features. Note that they only reported results with up to 64 bits, so their curves are incomplete here. Surprisingly, both VDSH and SDH work significantly better than the competitors. VDSH is consistently the best, delivering robust performance across all code lengths. FastHash tends to have good MAP performance, however, its precision within Hamming radius 2 drops drastically with longer hash codes, which is indicative of its inability to form compact clusters in the hash code space.

Evidently, the robust behavior suggests that the hash codes generated by VDSH in testing are sufficiently well clustered that data samples from the same class are mapped to nearby hash codes. 
We verify our conjecture by comparing VDSH, SDH and FastHash on CIFAR-10: (1) we visualize the hash codes with 64 bits of all the test images using t-SNE in Fig. \ref{fig:Visualization-CIFAR-10}, and (2) we directly report the precision and recall \wrt different code lengths with Hamming radius equal to 0, 1, and 2, respectively, in Fig. \ref{fig:hamming}.

As we see in Fig. \ref{fig:Visualization-CIFAR-10}, with different features VDSH forms cleaner clusters relative to SDH, suggesting good retrieval performance\footnote{Note that (a) appears to have fewer points than (b), but in fact there are the same number of points in both plots and many of the bit codes for the same classes collapse to the same 2D points in (a). Similarly we see this in (c) and (d) as well.}. This visual observation implies that, for VDSH, during testing a query image typically falls into or near the cluster belonging to its ground-truth class. This leads to Hamming distance being relatively small for the archival data within the same class than for other methods.

We next plot performance for decreasing Hamming radius in
Fig. \ref{fig:hamming}. VDSH appears to be robust and does not suffer performance degradation with decreasing radius. In contrast the performance of SDH and FastHash varies significantly and they both achieve the best result within Hamming radius 2. This finding further strengthens our view that VDSH is capable of learning compactly clustered hash codes across different code lengths (see also  Fig. \ref{fig:Visualization-CIFAR-10}). 

\begin{figure}[t]
\begin{minipage}[b]{.42\linewidth}
 \begin{center}
 \centerline{\includegraphics[width=.9\columnwidth]{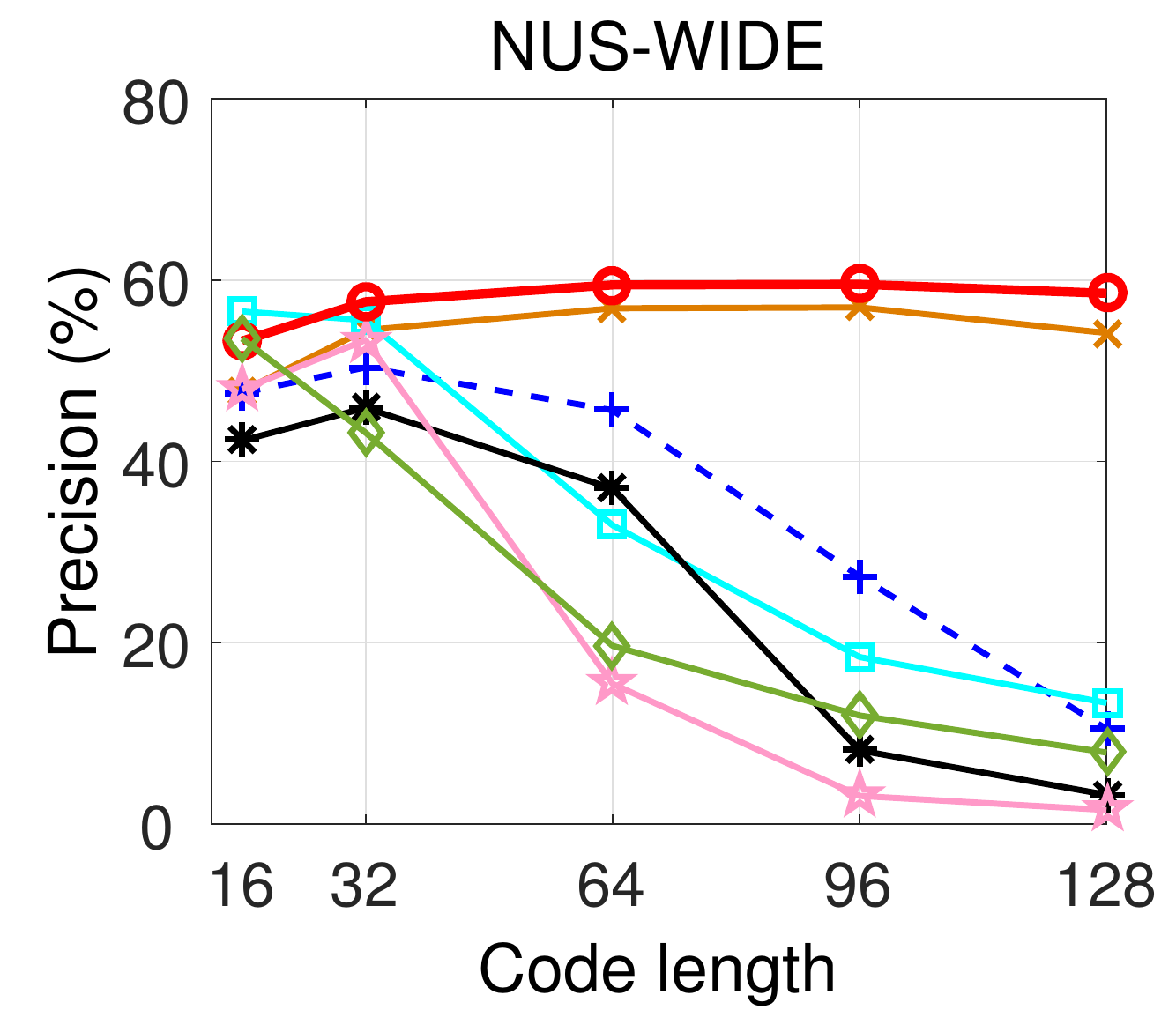}}
\centerline{\footnotesize{(a)}} 
 \end{center}
\end{minipage}
\begin{minipage}[b]{.55\linewidth}
\begin{center}
\centerline{\includegraphics[width=.9\columnwidth]{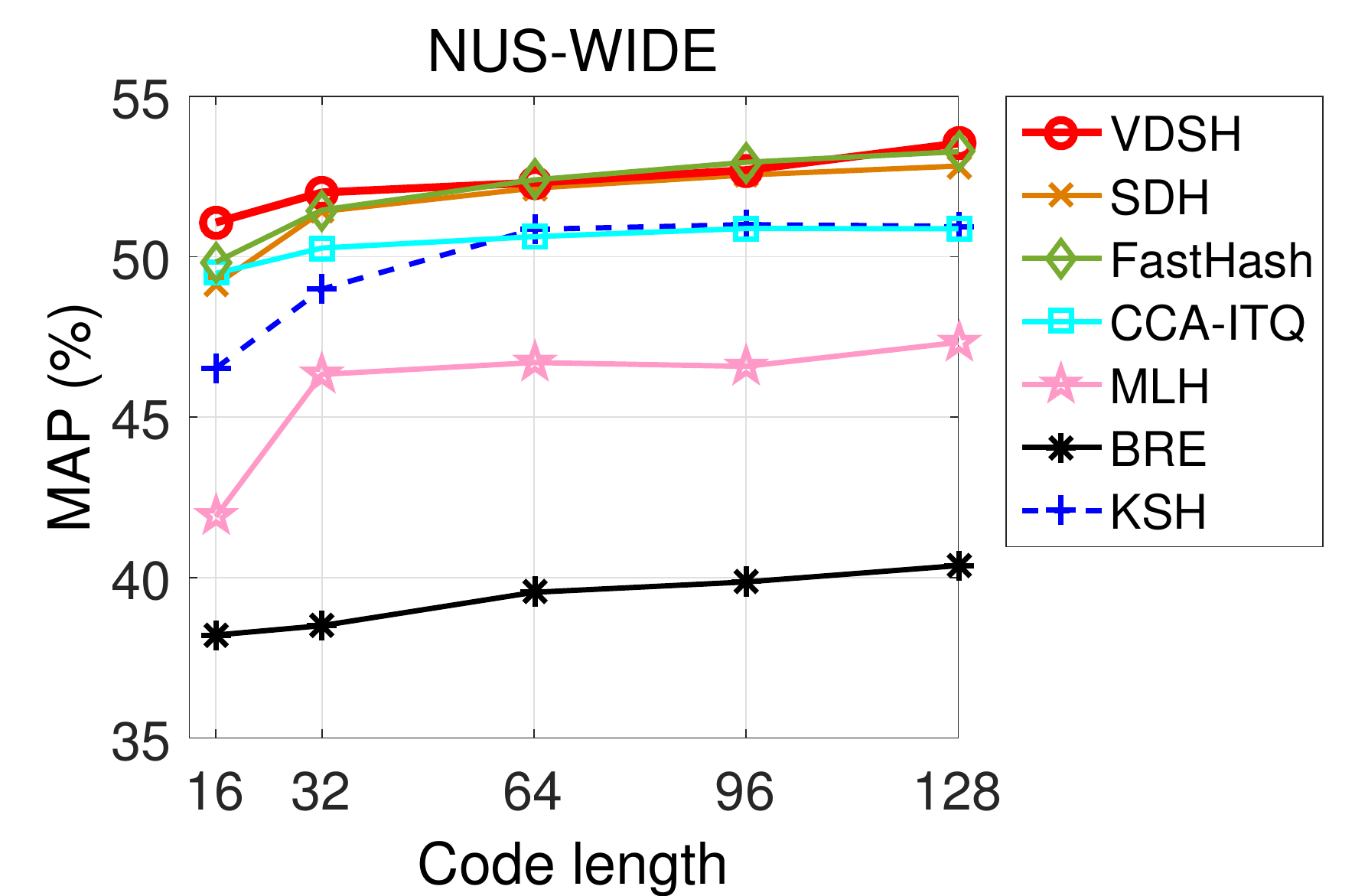}}
\centerline{\footnotesize{(b)}} 
\end{center} 
\end{minipage}
\vspace{-3mm}
\caption{\footnotesize{Precision and MAP comparison on NUS-WIDE with Hamming radius equal to 2. The features used here are the bag-of-words feature vectors provided by the dataset.
}}
\label{fig:NUS-WIDE}
\vspace{-2mm}
\end{figure}

Finally we test VDSH on NUS-WIDE using a network with 32 hidden layers and 128 nodes per layer. It takes less than 5 minutes for training, and 31.4 milliseconds per sample for hash code generation to retrieve a 190K-sample database. Performance comparisons are depicted in Fig.~\ref{fig:NUS-WIDE}. For CCA-ITQ, SDH and VDSH, the entire retrieval database is used for training. For the other methods, their huge memory requirements limit us to randomly sample 5K images for training. Here VDSH consistently achieves the best. But the performance gap between VDSH and SDH is not as significant as those in Fig. \ref{fig:PreRecMAP_mnist_cifar}. 
We hypothesize that this could be due to the fact that this is a multi-label dataset. Since we define two images to be neighbors when they share one common label, about 36 percent of the image pairs in this dataset are defined to be neighbors, compared with around 5 percent for a single label dataset of the same scale. The feature spaces of different classes (\ie concepts) thus tend to have large overlap. Our VDSH network could get confused by the same training samples that belong to different classes and thus unable to generate very effective hash codes. Another possibility is that the performance using the provided bag-of-words features may be already saturated.

In addition, we compare our method with others on ILSVRC2012 \cite{ILSVRC15}. Same as SDH \cite{shen2015supervised}, we randomly select 10 images for each of 1K classes in the training dataset from ILSVRC2012 to create a 10K-image training set to train different hashing methods, and utilize the entire 50K-image validation dataset in ILSVRC2012 as the query set. We first extract 4096-dim features using the vgg-f model. Then we compare our VDSH with SDH and FastHash based on 64-bit hash codes within Hamming radius 2, and here are the results (method, precision, recall): (VDSH, 7.73\%, 4.82\%), (SDH, 2.67\%, 0.96\%), (FastHash, 0.29\%, 0.61\%). Clearly, our method is still remarkably better than the state-of-the-art for supervised hashing.

\section{Conclusion}
In this paper, we propose a very deep supervised hashing (VDSH) algorithm to learn hash codes by training very deep neural networks. Our VDSH utilizes the outputs of DNNs to generate hash codes by rounding. For computational efficiency we formulate the training of VDSH as an $\ell_2$ norm regularized least square problem and propose a novel ADMM based training algorithm which can overcome the issues such as vanishing gradients in the traditional backprop algorithm by decomposing network-wide training into multiple independent layer-wise local updates. We discuss the empirical convergence and computational complexity of our training algorithm, and illustrate the weights learned by the networks. We conduct comprehensive experiments to compare VDSH with other (deep) supervised hashing methods on three benchmark datasets (\ie MNIST, CIFAR-10, and NUS-WIDE), and VDSH outperforms the state-of-the-art significantly.


\section*{Acknowledgement}
We thank the anonymous reviewers for their very useful comments. This material is based upon work supported in part by the U.S. Department of Homeland Security, Science and Technology Directorate, Office of University Programs, under Grant Award 2013-ST-061-ED0001, by ONR Grant 50202168 and US AF contract FA8650-14-C-1728. The views and conclusions contained in this document are those of the authors and should not be interpreted as necessarily representing the social policies, either expressed or implied, of the U.S. DHS, ONR or AF.

{\footnotesize
\bibliographystyle{ieee}
\bibliography{egbib}
}

\end{document}